\documentclass{article}

\usepackage[a4paper, total={6.5in, 9in}]{geometry}

\usepackage{url}
\usepackage{lmodern}
\usepackage[utf8]{inputenc}
\usepackage{placeins}
\usepackage{pgfplots}
\usepackage{subcaption}
\DeclareUnicodeCharacter{2212}{−}
\usepgfplotslibrary{groupplots,dateplot}
\usetikzlibrary{patterns,shapes.arrows}
\pgfplotsset{compat=newest}

\usepackage{graphicx}
\usepackage{authblk}
\usepackage{xcolor}
\usepackage[colorlinks = true,
            linkcolor = red,
            urlcolor  = blue,
            citecolor = blue,
            anchorcolor = red]{hyperref}
            
\usepackage{algorithm}
\usepackage{algpseudocode}
\usepackage{natbib}

\usepackage{amssymb}
\usepackage{amsthm}
\usepackage{thmtools} 
\usepackage{thm-restate}
\newtheorem{theorem}{Theorem}
\newtheorem{corollary}{Corollary}
\newtheorem{defn}{Definition}
\newtheorem{lemma}{Lemma}

\newcommand{\EE}{\mathbb{E}}

\newcommand{\bin}{\text{Bin}}

\newcommand{\arxiv}[1]{#1}
\newcommand{\conf}[1]{}

\usepackage{amsmath,amsfonts,bm}

\def\eqref#1{equation~\ref{#1}}

\def\1{\bm{1}}

\DeclareMathAlphabet{\mathsfit}{\encodingdefault}{\sfdefault}{m}{sl}
\SetMathAlphabet{\mathsfit}{bold}{\encodingdefault}{\sfdefault}{bx}{n}

\def\gT{{\mathcal{T}}}

\def\gX{{\mathcal{X}}}

\def\sR{{\mathbb{R}}}

\begin{document}

\title{Rate of Model Collapse in Recursive Training}

\author[1]{Ananda Theertha Suresh \footnote{\texttt{theertha@google.com}}}
\author[2]{Andrew Thangaraj}
\author[2]{Aditya Nanda Kishore Khandavally}
\affil[1]{Google Research, New York}
\affil[2]{Indian Institute of Technology Madras}

\maketitle

\begin{abstract}
Given the ease of creating synthetic data from machine learning models, new models can be potentially trained on synthetic data generated by previous models. This recursive training process raises concerns about the long-term impact on model quality. As models are recursively trained on generated data from previous rounds, their ability to capture the nuances of the original human-generated data may degrade. This is often referred to as \emph{model collapse}. In this work, we ask how fast model collapse occurs for some well-studied distribution families under maximum likelihood (ML or near ML) estimation during recursive training. Surprisingly, even for fundamental distributions such as discrete and Gaussian distributions, the exact rate of model collapse is unknown. In this work, we theoretically characterize the rate of collapse in these fundamental settings and complement it with experimental evaluations. Our results show that for discrete distributions, the time to forget a word is approximately linearly dependent on the number of times it occurred in the original corpus, and for Gaussian models, the standard deviation reduces to zero roughly at $n$ iterations, where $n$ is the number of samples at each iteration. Both of these findings imply that model forgetting, at least in these simple distributions under near ML estimation with many samples, takes a long time.
\end{abstract}

\section{Introduction}

\begin{figure*}[htb]
\begin{center}
\conf{ \includegraphics[scale=0.2]{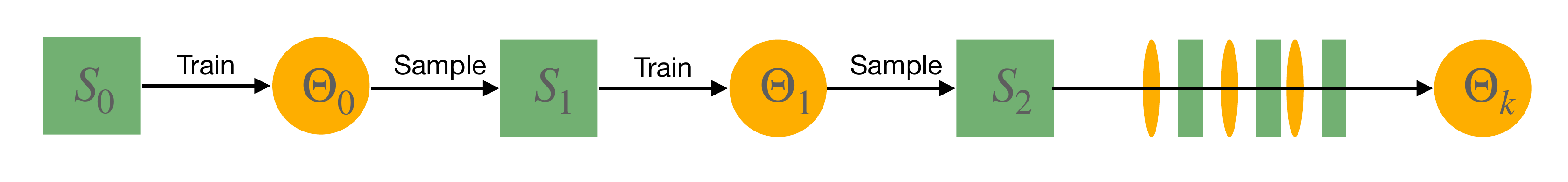}
}
\arxiv{ \includegraphics[scale=0.25]{recursive.pdf}
}
\caption{Simplified recursive training procedure of \cite{shumailov2023curse}. The underlying dataset is denoted by $S_0$. The model parameter $\Theta_k$ is trained based on dataset $S_{k}$ and sample $S_{k+1}$ is obtained by sampling from model with parameter $\Theta_k$.}
\label{fig:recursive}
\end{center}
\end{figure*}

Machine learning models are trained using datasets that, in most cases, are originally human generated. However, with the recent advance of generative models \citep{brown2020language,  chowdhery2022palm, thoppilan2022lamda, touvron2023llama}, it is possible today for machine learning models to generate human-like data.
This has led to the proliferation of machine-generated text and images in all forms of digital content. Hence, in the next iteration of model training, if one collects a large corpus of data such as text or images, a part of it would be human generated and a part  would be machine generated. Suppose we repeat this \emph{recursive training} process several times, i.e. we repeatedly train models on data generated from previous iteration of models. Then, it is reasonable to suspect that the model quality might degrade over time. Motivated by this concern, \cite{shumailov2023curse, shumailov2024ai} studied the problem of recursive training and showed that the tail of the original dataset disappears over multiple rounds of recursive training. In this work, we take a step further and ask \emph{how fast does the model quality degrade in recursive training?} In order to answer this question, we study the simplified recursive training procedure of \cite{shumailov2023curse} where models at current iteration are trained solely based on data generated by models at the previous iteration. Figure~\ref{fig:recursive} demonstrates the iterative training procedure. 

Let $P_{\theta}$ denote a distribution over a domain $\gX$ parameterized by a vector $\theta = (\theta_1,\theta_2,\ldots, \theta_m)$. For example, in the case of Gaussian distributions, $\theta = (\mu, \sigma^2)$. Let $S_0 = (X_{0, 1}, X_{0, 2}, \ldots, X_{0,n})$ denote the original dataset of $n$ samples, which can come from any arbitrary distribution.  Let $\Theta_k=(\Theta_{k,1},\Theta_{k,2},\ldots, \Theta_{k,m})$  denote the vector of model parameters at iteration $k$, and let $P_{\Theta_k}$ denote the corresponding sampling distribution. For each $k \geq 0$, the model $P_{\Theta_k}$ is trained based on dataset $S_{k} = (X_{k, 1}, X_{k, 2}, \ldots, X_{k,n})$, and then $P_{\Theta_k}$ is sampled to create a dataset $S_{k+1} = (X_{k+1, 1}, X_{k+1, 2}, \ldots, X_{k+1,n})$ for the next iteration. We have the recursion for $k=0,1,\ldots$:
\begin{equation}
\begin{aligned}
\Theta_{k} &= \widehat{\theta}(X_{k,1},\ldots,X_{k,n}) \quad \text{(train)}, \\
S_{k{+}1}{=}(&X_{k{+}1,1},\ldots,X_{k{+}1,n}) \sim \text{iid }P_{\Theta_k}\ \text{(sample)},
\end{aligned}
\label{eq:rectrain2}
\end{equation}
where $\widehat{\theta}$ denotes an estimator for the parameter vector from $n$ iid samples in each iteration. 

\begin{figure*}[t]
    \centering
    \begin{subfigure}{0.665\textwidth}
        \centering
        \includegraphics[width=1\textwidth]{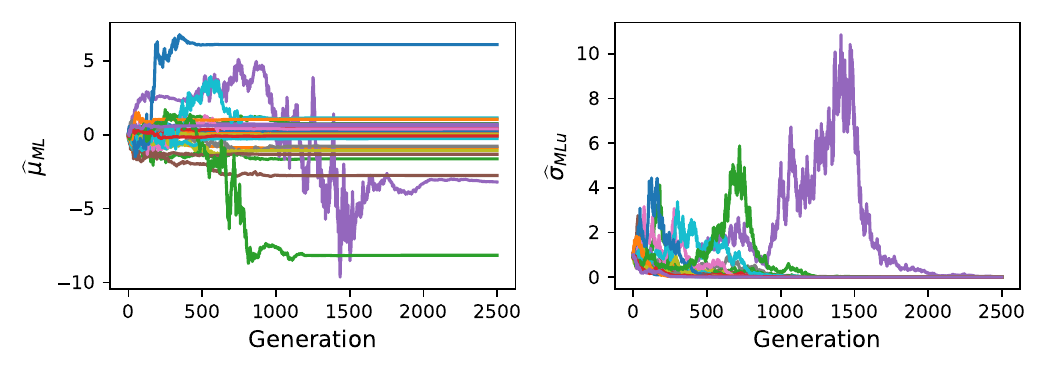}
        \caption{Gaussian model ($\mu_0 = 0$, $\sigma_0 = 1$)}
    \end{subfigure}%
    \hfill
    \begin{subfigure}{0.335\textwidth} 
        \centering
        \includegraphics[width=1\textwidth]{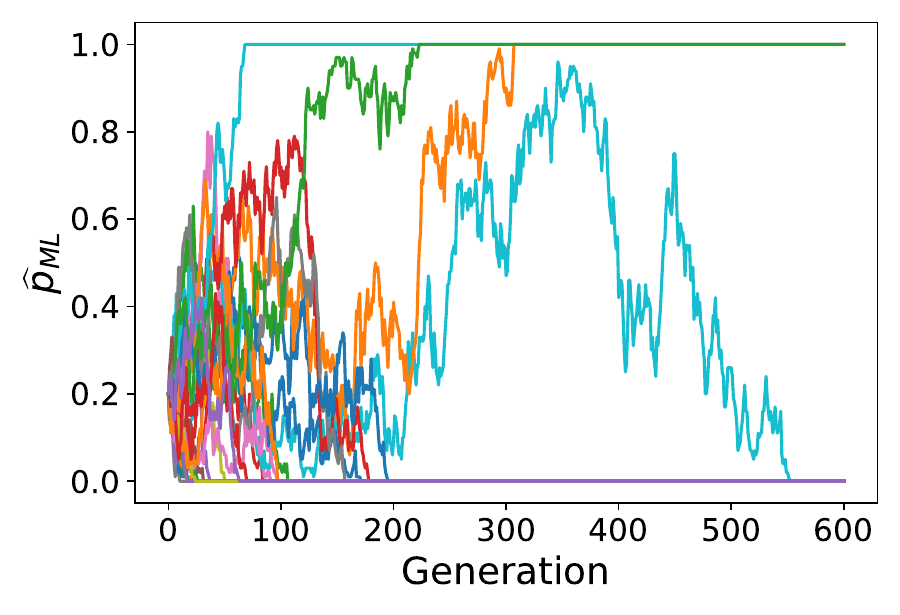}
        \caption{Bernoulli model ($p_0 = 0.2$)}
    \end{subfigure}
    \caption{Multiple trajectories of recursive training across generations in the Gaussian and Bernoulli models, with $n=100$ samples per generation. }
    \label{fig:trajectories}
    \end{figure*}

Our goal is to study the behaviour of the random process $\{\Theta_k:k=0,1,2,\ldots\}$, and observe convergence properties of the parameters in $\Theta_k$ as $k\to\infty$. If model quality is not degraded over recursive training, sample paths of $\Theta_k$ will be ergodic and cover the entire space of possible parameters. However, if any of the parameters are tending to zero or some other trivial value, we have an instance of so-called \emph{model collapse} in recursive training \citep{shumailov2023curse}. In such cases, we attempt to characterise the rate (as a function of $k$) at which the collapsing parameter tends to the trivial value. As an illustrative example, we plot $25$ different trajectories of recursive training starting with $N(0, 1)$ in Figure~\ref{fig:trajectories}(a). 
Observe that the standard deviation tends to zero always and the means tend to different values. Figure \ref{fig:trajectories}(b) \footnote{Code can be found at \url{www.github.com/berserank/rate-of-model-collapse/} } shows 25 trajectories for recursive training of Bernoulli samples. When started with Bernoulli$(0.2)$, the estimated Bernoulli parameter tends to zero or one in each trajectory. 

The recursive training iteration in \eqref{eq:rectrain2} can be termed a stochastic recursion, which is a broad topic with applications ranging from dynamical systems and optimization to reinforcement learning \citep{borkar2022stochastic, Diaconis99} (see Appendix~\ref{sec:app_related} for more details). 
While the nature of the stochastic recursion in \eqref{eq:rectrain2} does not match exactly with previously studied cases, a lot of underlying ideas are useful. Firstly, there is a Markov structure inherent in \eqref{eq:rectrain2}. Given $S_k$, the future models do not depend on the $S_{k-1}$ and the past. Secondly, under suitable unbiasedness conditions on $\widehat{\theta}$, we can expect martingale properties \citep{williams1991probability} to hold for certain functionals of $\Theta_k$. Finally, for convergence in stochastic recursions, an average \emph{contraction} property in each step of the recursion has been seen to be very useful to prove convergence bounds. So, one viable approach is to prove contraction properties for the expectation of a suitable function of parameters in $\Theta_{k}$ and ultimately derive convergence rates.

\subsection{Our contributions}
In this article, we primarily consider two types of models. Motivated by the language modelling applications, we first study learning a discrete distribution $P_{\theta}$ with probability mass function $\theta=(\theta_{1},\ldots,\theta_{m})$ over a finite alphabet $\{1,2,\ldots,m\}$. 
In the terminology of language models, $\theta$ corresponds to the unigram distribution. Under the iterations in \eqref{eq:rectrain2}, if $\Theta_{k,i}\to0$ as $k\to\infty$, the model forgets the symbol $i$ under recursive training, and this is an instance of model collapse. 

Secondly, motivated by image generation tasks, we study one-dimensional Gaussian and Gaussian mixture distributions where $\theta=(\mu,\sigma)$. In the Gaussian case, we have $P_{\theta}=N(\mu,\sigma^2)$, which denotes the Gaussian distribution with mean $\mu$ and variance $\sigma^2$. In the Gaussian mixture case, we let $P_{\theta}=\frac{1}{2}N(-\mu,\sigma^2)+\frac{1}{2}N(\mu,\sigma^2)$. In these two cases, if the variance parameter tends to zero under recursive training, we have model collapse.

During our analysis, we also characterize rates to model collapse for several models such as Bernoulli and Poisson models, which might be interesting on their own. Our results show the following interesting behaviors:
\begin{itemize}
\item For discrete distributions, picking $\widehat{\theta}$ as the Maximum Likelihood (ML) estimator, we show that $\text{Pr}(\Theta_{k,i}\ne0)\geq 1 - \exp\left( - \frac{\lambda_i}{k} \right)$,
 where $\lambda_i$ is the 
 number of $i$'s in the original dataset $S_0$. So, the probability-rate of forgetting symbol $i$ under recursive training falls as $1/k$, which is quite a slow rate of fall. If today's complex language models move closer to near ML estimation, it may take a very large number of recursive training rounds before relatively frequent words and phrases are actually forgotten. 
\item For the one-dimensional Gaussian model with $\widehat{\theta}$ as the ML estimator, we show that variance tends to zero and mean tends to a constant as $k\to\infty$. 
Our convergence proof shows that $\text{Pr}(\text{variance at iteration }k>\epsilon^2)$ falls roughly as $\frac{1}{\epsilon} e^{-k/(4n)}$. Under this bound, we get that $k$ for which variance likely goes below $\epsilon^2$ grows roughly linearly with $n$. Since $n$ is very large in practice, model collapse could possibly be very slow, assuming estimators are close to optimal.
\item For one-dimensional Gaussian mixtures with $\widehat{\theta}$ as an approximate version of the joint ML estimator for $\mu$ and $\sigma$, we show that variance tends to zero and mean tends to a constant as $k\to\infty$ proving model collapse. 
The rate of collapse is similar to Gaussian and is $\frac{1}{\epsilon} e^{-k/(4n)}$.
\end{itemize}
While all the distributions we study are almost toy-like versions of their practically useful counterparts, the exact rate of forgetting or model collapse under recursive training is not previously known for any of these models to the best of our knowledge. 
\subsection{Related works}

The problem of recursive training has garnered significant interest in the last few years. Due to space constraints, we refer readers to Appendix~\ref{sec:app_related} for  related works and focus on works that are most relevant to ours here. For the Gaussian distribution estimation, the most relevant work to ours are by \cite{alemohammad2023self, shumailov2024ai, bertrand2023stability}. 
\cite{alemohammad2023self, shumailov2024ai}  showed that the variance of the resulting Gaussian tends to zero and \cite{bertrand2023stability} demonstrated an exponential decay of variance, but does not give a closed form result. Their bound can only be evaluated through simulations. In comparison, we provide an explicit and simple expression for the bound. For discrete distributions, the closest work to ours is \cite{seddik2024bad} who studied the problem of model collapse for theoretical distributions. As we show later both theoretically and empirically, the proposed results are stronger than bounds of \citet[Equation 7]{seddik2024bad}. 

 In the next section, we describe the problem setting and in the followup sections,  we state our main theoretical results. Due to space constraints all the proofs are provided in the appendix.

\section{Problem setting}
\label{sec:prob}

A recursive training process is specified by the sampling distribution $P_\theta$ with the associated parameter vector $\theta$, the initial parameter $\theta_0$, the number of samples in each round $n$ and the estimator $\widehat{\theta}$. It is clear from ~\eqref{eq:rectrain2} that $\{\Theta_k\}$ is a homogeneous Markov chain. We will be interested in the convergence properties of $\{\Theta_k\}$.  To keep the notation clear, we will denote random variables by capital letters and non-random values by small letters.

\subsection{Model collapse in recursive training}
We are now ready to formally define model collapse in recursive training as studied in this paper. Suppose the parameter space $\gT$ is the set of all possible values of the parameter vector $\theta=(\theta_1,\theta_2,\ldots, \theta_m)$. We will let $\gT_i$ denote the set of possible values of $\theta_i$. The richness of a model may be characterized by the size or dimension of the entire parameter space $|\gT|$ and/or that of an individual parameter $\gT_i$.

 In the recursive training process, $\Theta_k$ takes values in the range of the estimator $\widehat{\theta}$. Denoting the range of any function $f$ as $R(f)$, the richness in the models under recursive training are seen to be clearly limited by the size of the range of the estimator $R(\widehat{\theta})$. For an individual parameter $\theta_i$, the range of the estimator is $R(\widehat{\theta}_i)$. While it seems natural to have $\gT_i=R(\widehat{\theta}_i)$, there are cases such as Bernoulli$(p)$ distribution with $p\in[0,1]$ where $R(\widehat{p})\in\{0,1/n,2/n,\ldots,1\}$ for $n$ samples. If the sample paths of $\{\Theta_k\}$ traverse through the entire range $R(\widehat{\theta})$ (with high probability), then the model quality can be expected to be retained in the recursive training process given an estimator $\widehat{\theta}$. We use the opposite of this scenario to precisely define model collapse. 
\begin{defn}
    Model collapse is said to occur in a recursive training process $\{\Theta_k\}$ (as defined in the introduction) if, for some $i$, $\Theta_{k,i}\to0$ or a set of small size when compared to $R(\widehat{\theta}_i)$ (finite in size, if $|R(\widehat{\theta}_i)|$ is infinite, or $o(|R(\widehat{\theta}_i)|)$, if $|R(\widehat{\theta}_i)|$ is finite).
\end{defn}

\subsection{Martingale property and convergence}
Suppose $f:\gT\to[0,\infty)$ is a non-negative, bounded functional on the parameter space. If the estimator $\widehat{\theta}$ and the induced conditional law on $\Theta_{k+1}$ given $\Theta_k$ are such that $\EE[f(\Theta_{k+1})\,\vert\,\Theta_k]=f(\Theta_k)$, then the process $\{f(\Theta_k)\}$ is a non-negative, bounded martingale under the filtration generated by $\Theta_k$. By convergence properties of such martingales \cite{williams1991probability}, every sample path of $f(\Theta_k)$ converges to some finite value. 

The condition $\EE[f(\Theta_{k+1})\,\vert\,\Theta_k = \theta] = f(\theta)$ essentially requires that the plugin estimator $f(\widehat{\theta}(X_{k,1},\ldots,X_{k,n}))$ for $f(\theta)$ is unbiased. Since unbiasedness is a common property, it is likely that many \emph{good} estimators designed to minimize some risk or maximize likelihood will satisfy the required martingale condition for some $f$. We demonstrate such functionals for several interesting examples of sampling distributions $P_{\theta}$ and estimators $\widehat{\theta}$ in our results. 

However, the martingale convergence property alone does not directly imply model collapse. While every sample path will converge to some finite value, these values may be different for different sample paths, and they may be infinite in number. So, the convergence to a set of finite size needs to be established explicitly to prove model collapse. 

\subsection{Rate of model collapse} 
For simplicity of description, we will consider the collapse of an individual non-negative real model parameter $\theta_i>0$ for a fixed $i$ under recursive training as $k\to\infty$. To show collapse, we will establish that the random variable $\Theta_{k,i}$ (that models the evolution of $\theta_i$) converges to $0$ almost surely.

Our approach will be to show lower and upper bounds on $\text{Pr}(\Theta_{k,i}>\epsilon)$ to establish its rate of fall with $k$. Further, we will show $\text{Pr}(\cup_{k\ge m}\Theta_{k,i}>\epsilon)$ falls with $m$ to establish almost sure convergence to $0$. 
\section{Standard recursive training processes}

Having defined the setting of model collapse in recursive training, we now consider 
standard recursive training processes where the sampling distribution $P_{\theta}$ is Bernoulli, Poisson or Gaussian and the estimator $\widehat{\theta}$ is the Maximum Likelihood (ML) estimator.
\subsection{Bernoulli model}

\begin{figure}[t]
    \centering
     \conf{\resizebox{\linewidth}{!}}
     \arxiv{\resizebox{0.8\linewidth}{!}}{\input{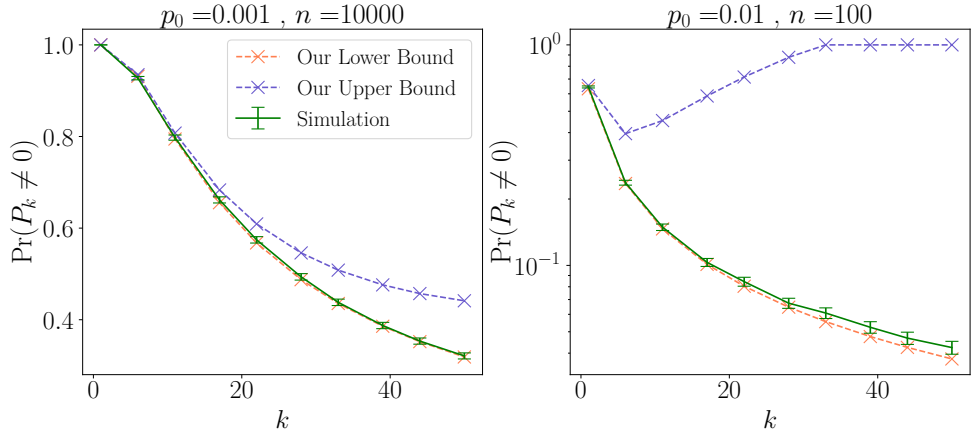}}
    \caption{Comparison of theoretical bounds and empirical observations for $\Pr(P_k \neq 0)$ at a particular generation in a Bernoulli recursive training process with $n$ samples at each generation. Results are shown for varying initial success probabilities $(p_0)$.}
    \label{fig:bernoulli}
\end{figure}

In this simple and instructional model, the sampling distribution is Bernoulli$(p)$ with a single real parameter $p\in[0,1]$. The Bernoulli recursive training process is denoted $\{P_k\}$ with $P_k$ distributed on $\gT=[0,1]$ and initialized as $P_0=p_0 = \frac{\sum_{i} X_{0,i}}{n}$. We will consider $n$ samples per round. Given $P_{k-1}=p$, the $n$ samples in the $k$-th round $X_{k,1},\ldots,X_{k,n}$ are iid Bernoulli$(p)$, and the ML estimate of $p$ from the $n$ samples is 
$$\widehat{p}_{ML}(X_{k,1},\ldots,X_{k,n})=\frac{X_{k,1}+\cdots+X_{k,n}}{n}.$$ 
By standard properties of the Bernoulli distribution, we have that,
$\text{given }P_{k-1}=p$,
\begin{equation*}
P_{k}=\widehat{p}_{ML}(X_{k,1},\ldots,X_{k,n})\sim\frac{1}{n}\text{Binomial}(n,p).
\end{equation*}
Hence, the Bernoulli recursive training process that we consider is described succinctly by the iteration
\begin{equation*}
    nP_{k}\sim \text{Binomial}(n,P_{k-1}).
\end{equation*}
It is readily seen that $nP_k\in\{0,1,\ldots,n\}$, $k=1,2,\ldots$, is a finite-state Markov chain with two absorbing states $0$ and $n$, while all the other states are transient. Clearly, every sample path of $P_k$ converges to either $0$ or $1$. This establishes model collapse for the Bernoulli recursive training process. By a standard computation of probabilities of absorption, it is possible to show that Pr$(\text{absorption to }0\,\vert\,nP_1=i)=\frac{n-i}{n}$ and that Pr$(\text{absorption to }0)=1-p_0$ (see Appendix~\ref{sec:bernoulliabsorption} for details). The following theorem determines bounds for the probability of absorption to $0$ at time $k$. 
\begin{restatable}{theorem}{bernoullitheorem}
\label{th:discrete_main}
    For the Bernoulli recursive training process defined above,
\begin{align*}
 \conf{ &} 1 - \left(1 - 3np^2_0 - 3 k p_0 \right) \exp(- n p_0 g_k )  \conf{\\
  &} \geq   \text{Pr}(P_k\ne0) \geq 1-\exp(-n p_0 g_k),
\end{align*}
where 
\begin{equation}
\label{eq:gk}
g_k \triangleq g^{\circ k}(\infty)\in[1/k,3/k]
\end{equation}
with $g(x)=1-e^{-x}$ and $g^{\circ k}$ denotes $g$ composed $k$ times. 
Furthermore, if $p_0 \leq \min \left( \frac{1}{6k}, \frac{1}{\sqrt{6n}} \right)$, then the lower bound is nearly tight:
\begin{equation*}
    \text{Pr}(P_k\ne0) \leq 1-0.5 \cdot \exp(-n p_0 g_k).
\end{equation*}
Similarly, bounds for $ \text{Pr}(P_k\ne 1)$ can be obtained by replacing $p_0$ with $1-p_0$ in the above set of equations.
\end{restatable}

We compare our results with empirical observations in Figure~\ref{fig:bernoulli} for different values of $p_0$. Our lower bound is the tightest and matches the empirical observation. 

\textbf{Comparison with \cite{seddik2024bad}:} For Bernoulli distributions, \citet[Theorem 1]{seddik2024bad} implies that $\Pr(P_k \notin \{0, 1\})$ lies in 
\begin{align*}
\left[ 4p_0(1-p_0) \left( 1 - \frac{1}{n}\right)^k,    2np_0(1-p_0)  \left( 1 - \frac{1}{n}\right)^k \right].
\end{align*}
Since our interest is on discrete distributions over large domains (Section~\ref{sec:discrete_main}), our focus is on regimes where $p_0$ is relatively small and hence it is unlikely that $P_k$ will reach one. However, by combining our bounds for $\Pr(P_k \neq 0)$ and $\Pr(P_k \neq 1)$, we can obtain bounds for $\Pr(P_k \notin \{0, 1\})$. In Appendix~\ref{sec:seddick}, we compare our results for $\Pr(P_k \notin \{0, 1\})$ to that of \cite{seddik2024bad} and demonstrate the tightness of our bounds.

\subsection{Poisson model}
\label{sec:poisson}
In this model, the sampling distribution is Poisson$(\lambda)$ with a single real parameter $\lambda>0$. The Poisson recursive training process is denoted $\{\Lambda_k\}$ with $\Lambda_k$ distributed on $\gT=[0,\infty)$ and initialized as $\Lambda_0=\lambda_0 = \frac{\sum_{i} X_{0,i}}{n}$. Given $\Lambda_{k-1}=\lambda$, the $n$ samples in the $k$-th round $X_{k,1},\ldots,X_{k,n}$ are iid Poisson$(\lambda)$, and the ML estimate of $\lambda$ from the $n$ samples is 
$$\widehat{\lambda}_{ML}(X_{k,1},\ldots,X_{k,n})=\frac{X_{k,1}+\cdots+X_{k,n}}{n}.$$ 
By standard properties of the Poisson distribution, we have that, $\text{given }\Lambda_{k-1}=\lambda$,
\begin{equation*}
\Lambda_{k}=\widehat{\lambda}_{ML}(X_{k,1},\ldots,X_{k,n})\sim\frac{1}{n}\text{Poisson}(n\lambda).
\end{equation*}
Therefore, the Poisson recursive training process is described by the iteration
$
    n\Lambda_{k}\sim \text{Poisson}(n\Lambda_{k-1}).
$
Letting $\tilde{\Lambda}_k=n\Lambda_k$, we have the iteration
\begin{equation*}
\label{eq:lambtilde}
    \tilde{\Lambda}_{k}\sim \text{Poisson}(\tilde{\Lambda}_{k-1}).
\end{equation*}
Since $\EE[\tilde{\Lambda}_{k+1}\,\vert\,\tilde{\Lambda}_k]=\tilde{\Lambda}_k$, $\tilde{\Lambda}_k$ is non-negative and $\EE[\tilde{\Lambda}_k]=n\lambda_0$ is bounded, we have that $\{\tilde{\Lambda}_k\}$ is a non-negative, bounded martingale. So, a sample path of $\tilde{\Lambda}_k$ converges to some finite value. Also, $\{\tilde{\Lambda}_k\}$, for $k=1,2,\ldots$, is a Markov chain with one absorbing state $0$. The following theorem explicitly determines the probability of absorption to $0$ at time $k$. 

\begin{figure}[t]
    \centering
    \conf{\resizebox{\linewidth}{!}}
     \arxiv{\resizebox{0.8\linewidth}{!}}{\input{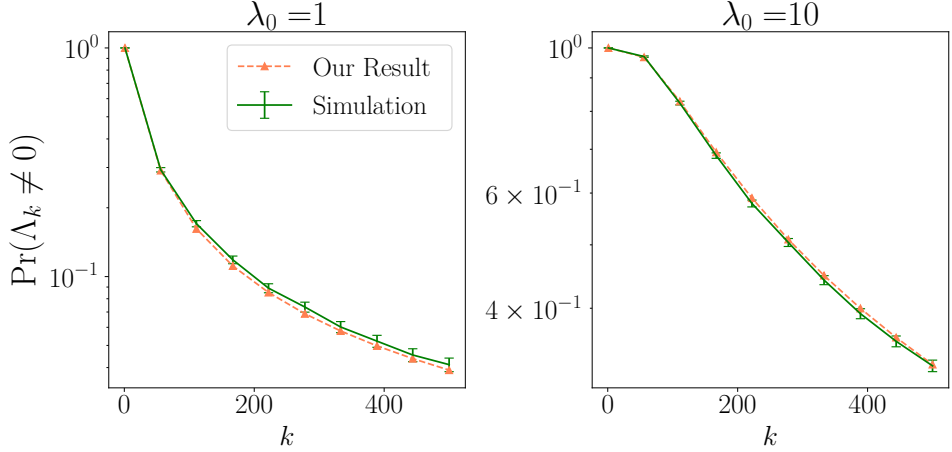}}
    \caption{Comparison of theoretical estimate and empirical observations for $\Pr(\Lambda_k \neq 0)$ at a particular generation in a Poisson recursive training process with $n = 10$ samples at each generation. Results are shown for varying initial arrival rate $\lambda_0$.}
    \label{fig:poisson}
\end{figure}

\begin{restatable}{theorem}{poissontheorem}
\label{thm:poisson}
For the Poisson recursive training process defined above,
\begin{equation*}
    \text{Pr}(\Lambda_k\ne0)=1-e^{-n\lambda_0 g_k}\in [1 - e^{-\frac{n\lambda_0}{k}},1 - e^{-\frac{3n\lambda_0}{k}}],
\end{equation*}
where $g_k$ is defined in~\eqref{eq:gk}. This implies that the parameter $\lambda$ in the Poisson$(\lambda)$ model collapses to $0$ almost surely under recursive training.
\end{restatable}
For the Poisson recursive training process, we have that the event $(\Lambda_m=0)$ is identical to the event $\cap_{k\ge m}(\Lambda_k=0)$. So, the above theorem implies that $\Lambda_k\to0$ almost surely.
We notice that $\text{Pr}(\Lambda_k\ne0)<\delta$ for $k>3n\lambda_0/\log\frac{1}{1-\delta}$. By increasing $n$, model collapse can be delayed linearly in Poisson recursive training. In Figure~\ref{fig:poisson}, we empirically demonstrate that the empirical estimate of $\text{Pr}(\Lambda_k\ne0)$ matches the result in Theorem~\ref{thm:poisson}.

\subsection{Gaussian model}
In this model, the sampling distribution is $N(\mu,\sigma^2)$ with two real parameters - mean $\mu$ and variance $\sigma^2\ge0$. The Gaussian recursive training process is denoted $\{(M_k,\Sigma_k)\}$ with $M_k$ distributed on $\sR$, $\Sigma_k$ on $[0,\infty)$ and  initialized as $M_0=\mu_0$, $\Sigma_0=\sigma_0$ (sample mean and standard deviation of initial samples). Given $M_{k-1}=\mu$, $\Sigma_{k-1}=\sigma$, the $n$ samples in the $k$-th round $X_{k,1},\ldots,X_{k,n}$ are iid $N(\mu,\sigma^2)$, and the ML estimate of $\mu$ and the bias-adjusted ML estimate of $\sigma^2$ from the $n$ samples are given by
\begin{align*}
    \widehat{\mu}_{ML}(X_{k,1},\ldots,X_{k,n}) &=\frac{X_{k,1}+\cdots+X_{k,n}}{n},\\
    \widehat{\sigma}^2_{MLu}(X_{k,1},\ldots,X_{k,n}) &=\frac{\sum^n_{i=1}(X_{k,i}-\widehat{\mu}_{ML})^2}{n-1}.
\end{align*}
By standard properties of Gaussian sample statistics, we have that, given $M_{k-1}=\mu, \Sigma_{k-1}=\sigma$,
\begin{equation*}
\begin{gathered}
M_{k}=\widehat{\mu}_{ML}(X_{k,1},\ldots,X_{k,n})\sim N\left(\mu,\frac{\sigma^2}{n}\right),\\
\text{is independent of}\\
\Sigma^2_{k}=\widehat{\sigma}^2_{MLu}(X_{k,1},\ldots,X_{k,n})\sim \sigma^2\frac{\chi^2_{n-1}}{n-1},    
\end{gathered}
\end{equation*}
where $\chi^2_{n-1}$ is the chi-squared distribution with $n-1$ degrees of freedom. So, the Gaussian recursive training process can be described by the recursion
\begin{equation*}
    M_{k+1}\sim N\left(M_k,\frac{\Sigma_k^2}{n}\right),\ \Sigma_{k+1}^2\sim \Sigma^2_k\frac{\chi^2_{n-1}}{n-1}.
\end{equation*}
We see that $\EE[\Sigma^2_{k+1}\,\vert\,\Sigma^2_k]=\Sigma^2_k$ (note that mean of $\chi^2_{n-1}$ is $n-1$) and $\Sigma^2_k$ is a martingale. Since $\EE[\Sigma_k^2]=\sigma^2$ and $\Sigma^2_k$ is non-negative, a sample path of $\{\Sigma^2_k\}$ converges to some finite value by martingale convergence theorems. The following theorem asserts that the sample paths of $\Sigma^2_k$ converge to $0$ implying that the variance parameter in the Gaussian recursive training process collapses to $0$.
\begin{theorem}
    For the Gaussian recursive training process defined above with $n\ge3$,
    \begin{align}
    \Pr(\Sigma_k>\epsilon)&\le \frac{\sigma_0}{\epsilon}\exp\left\{-\frac{k}{4n-1}\right\}.\nonumber
    \end{align}
    Hence, the parameter $\sigma$ in the $N(\mu,\sigma^2)$ model collapses to $0$ almost surely under recursive training. 
    \label{thm:Gaussian}
\end{theorem}

\begin{figure}[t]
    \centering
     \conf{\resizebox{0.75\linewidth}{!}}
     \arxiv{\resizebox{0.5\linewidth}{!}}
{\input{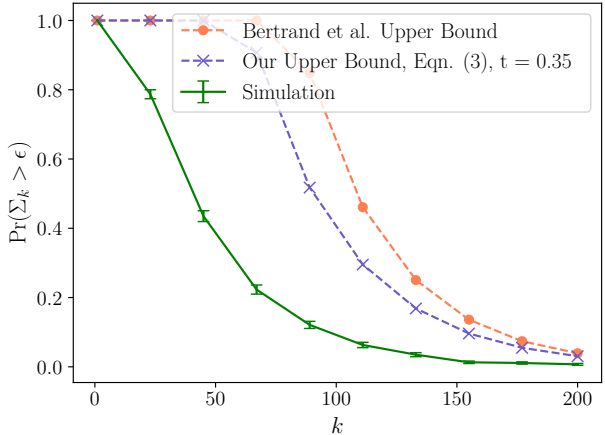}}
    \caption{Comparison of theoretical upper bounds with empirical observations of  $\text{Pr}(\Sigma_k>\epsilon)$ in Gaussian recursive training process $(\epsilon = 10^{-1}, n = 10)$.}
    \label{fig:Gaussianprobbounds}
\end{figure}

A proof using manipulations of the moments of chi squared distributions can be found in Appendix \ref{sec:Gaussian}. In the proof, we show that
\begin{equation}
\text{Pr}(\Sigma_k\ge\epsilon)\le \min_{t\in[0,1]}\left(\frac{\sigma^2_0}{\epsilon^2}\right)^t\left(\frac{\Gamma(t+(n-1)/2)}{((n-1)/2)^t\Gamma((n-1)/2)}\right)^k,
\label{eq:GaussianGammabound}
\end{equation}
which is upper bounded for $t=1/2$ by the bound in Theorem \ref{thm:Gaussian}. The proof and the explicit and simple expression for the bound are significantly different when compared to the bound in \cite{bertrand2023stability}, which needed to be evaluated through simulations. Our bound is illustrated in Figure \ref{fig:Gaussianprobbounds} and compared to the bound in \cite{bertrand2023stability} evaluated by simulations. We see that our upper bound is noticeably better for a wide range of $k$. We remark that the same result as Theorem \ref{thm:Gaussian} holds even when the ML estimate for $\sigma^2$, $\widehat{\sigma}^2_{ML}=\frac{n-1}{n}\widehat{\sigma}^2_{MLu}< \widehat{\sigma}^2_{MLu}$, is used instead of $\widehat{\sigma}^2_{MLu}$.

Using the bound of Theorem \ref{thm:Gaussian} and the union bound,
\begin{align}
  \conf{  & }\text{Pr}\bigg(\bigcup\limits_{k\ge m}\Sigma_k>\epsilon\bigg)\le \sum_{k=m}^{\infty} \frac{\sigma_0}{\epsilon}\exp\left\{-\frac{k}{4n-1}\right\}
  \conf{\nonumber\\
    &\qquad\qquad }=\frac{\sigma_0/\epsilon}{1-\exp\left(-\frac{1}{4n-1}\right)}\exp\left(-\frac{m}{4n-1}\right).\nonumber
\end{align}
This shows that $\Sigma_k\to0$ almost surely.

Though $\{M_k\}$ is a martingale, it is not apparent that $\EE[|M_k|]$ is bounded and we cannot immediately use martingale convergence to argue that sample paths of $M_k$ will converge to a constant. However, since $\Sigma^2_k\to0$ almost surely and $M_{k+1}-M_k\sim N(0,\Sigma^2_k/n)$, we see that every sample path of $\{M_k\}$ converges to some finite value. Further, $\text{Pr}(\Sigma_k>\epsilon)<\delta$ for 
\begin{equation*}
    k > (4n-1)\log \frac{\sigma_0}{\epsilon\delta}.
\end{equation*}
The lower bound above scales linearly with $n$ and as $\log(1/\epsilon)$ in $\epsilon$. 
\section{Applications}
We will now present results for recursive training in models that are closer to applications. The standard models considered earlier will be of use here.

\subsection{Discrete distributions}
\label{sec:discrete_main}
\begin{figure}[t]
    \centering
        \conf{\resizebox{0.9\linewidth}{!}}
     \arxiv{\resizebox{0.5\linewidth}{!}}{\input{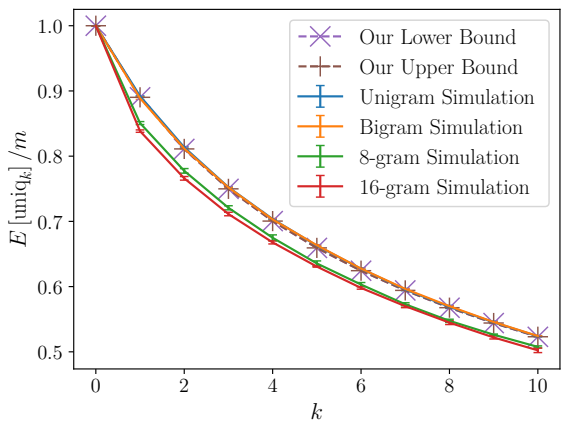}}
    \caption{Fraction of survived alphabet $\EE\left[\text{uniq}_k \right]/m$, over generations with recursive training in various n-gram language models. $n=1M$ tokens are generated at every generation. Error bars represent 1 standard deviation across 5 trajectories.}
    \label{fig:ngram_sampling}
\end{figure}

\begin{figure}
    \centering
     \conf{\resizebox{\linewidth}{!}}
     \arxiv{\resizebox{0.75\linewidth}{!}}{\input{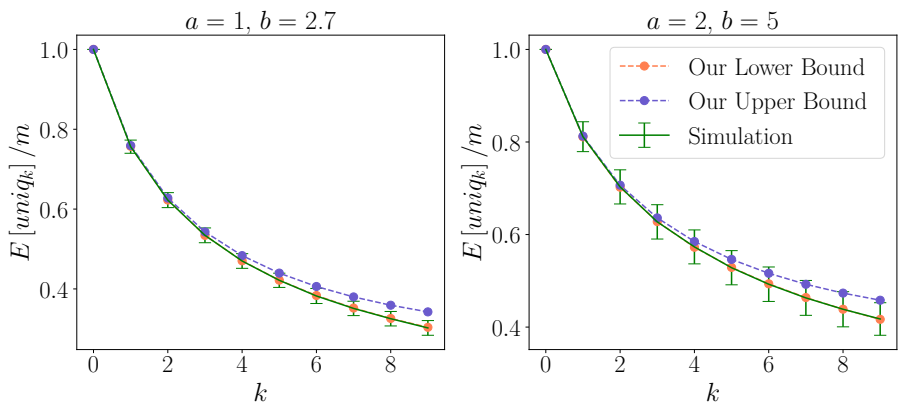}}
    \caption{Fraction of survived alphabet $\EE\left[\text{uniq}_k \right]/m$, over generations with recursive training following a Zipfian distribution (using various parameters $a$ and $b$) initially, for an alphabet size of $m = 1000$. At each generation, $n = 1000$ samples are generated. Error bars represent 1 standard deviation across 50 trajectories.}
    \label{fig:zipf}
\end{figure}

The sampling distribution is a discrete distribution $P_\theta$ on the alphabet $\{1,\ldots,m\}$ and the parameter vector $\theta=(\theta_1,\ldots,\theta_m)$ is the probability mass function. The recursive training process is denoted $\{\Theta_k\}$ with $\Theta_k$ distributed on the probability simplex in $\sR^m$ and initialized at $\Theta_0=\theta_0=(\theta_{0,1},\ldots,\theta_{0,m})$ based on $S_0$. Given $\Theta_k=\theta$, the $n$ samples in the $k$-th round $X_{k,1},\ldots,X_{k,n}$ are iid $\theta$, and the ML estimate of $\theta_i$ is 
$$\widehat{\theta}_{i,ML}(X_{k,1},\ldots,X_{k,n})=\frac{\sum^n_{j=1}I(X_{k,j}=i)}{n},$$ 
where $I(A)$ is the indicator for event $A$. As seen earlier, $\text{given }\Theta_{k-1,i}=\theta_i,$
\begin{equation*}
\Theta_{k,i}=\widehat{\theta}_{i,ML}(X_{k,1},\ldots,X_{k,n})\sim\frac{1}{n}\text{Binomial}(n,\theta_i).
\end{equation*}
So, in this case, we get the Bernoulli recursive training process iteration
\begin{equation*}
    n\Theta_{k+1,i}\sim \text{Binomial}(n,\Theta_{k,i}).
\end{equation*}
Hence for each symbol $i$, the probability that the symbol does not appear after $k$ iterations is given by Theorem~\ref{th:discrete_main}. We can use this to bound the number of symbols that appears after $k$ iterations.
\begin{corollary}
Let $\text{uniq}_k= \sum^m_{i=1}I(\widehat{\Theta}_{k,i} \neq 0)$ denote that number of distinct symbols after $k$ steps in the discrete distribution recursive training process. Then
\begin{align*}
\conf{&} \sum^m_{i=1} \left(1 - \exp(- n p_{0,i} g_k )\max\left(1 - 3np^2_{0,i} - 3 k p_{0,i}, 0 \right) \right)\conf{ \\
& }\geq \EE\left[\text{uniq}_k \right] \geq \sum^m_{i=1} \left(1 - \exp(-n  p_{0,i}  g_k )\right) ,
\end{align*}
where $g_k$ is defined in~\eqref{eq:gk}.
\end{corollary}
In Figure~\ref{fig:ngram_sampling}, we plot the estimate for number of distinct tokens at each generation for recursive n-gram language model sampling using Wikitext-2 corpus \citep{merity2016pointer}. We select $n = 1M$. Even though our theoretical approach is only valid for the unigram model, we note that our bounds hold tight even for higher order $n$-gram models. We also note that \cite{seddik2024bad} only provides guarantees for total model collapse, which in this scenario requires $O(n) = O(1M)$ iterations, whereas our results provide fine-grained guarantees on the number of distinct symbols. 
If Figure~\ref{fig:zipf}, we also plot the tightness of our upper and lower bounds for Zipfian distributions.

The above analysis assumes multinomial sampling where the number of samples is fixed at each iteration. Another way to sample from a discrete distribution is Poisson sampling, where at each round the number of samples is a Poisson random variable with mean $n$. We provide guarantees for Poisson sampling in Appendix~\ref{sec:poisson_discrete}.

\subsection{Gaussian mixture model (GMM)}
In the Gaussian mixture model (GMM) that we consider, the sampling distribution is $\frac{1}{2}N(-\mu,\sigma^2)+\frac{1}{2}N(\mu,\sigma^2)$ with two non-negative real parameters - mean $\mu\ge0$ and variance $\sigma^2\ge0$. The GMM recursive training process $\{(M_k,\Sigma_k)\}$ with $M_k$ and $\Sigma_k$ distributed on $[0,\infty)$ is initialized as $M_0=\mu_0$, $\Sigma_0=\sigma_0$ based on $S_0$. Given $M_k=\mu$, $\Sigma_k=\sigma$, the $n$ samples in the $k$-th round $X_{k,1},\ldots,X_{k,n}$ are iid $\frac{1}{2}N(-\mu,\sigma^2)+\frac{1}{2}N(\mu,\sigma^2)$. 

To estimate $\mu$ and $\sigma$, we introduce a new parameter $\alpha$ that can be viewed as noise to signal ratio. Lemma \ref{lem:gmmjointml} provides the following implicit equations whose solution gives the joint ML estimates with $n$ samples. 
\begin{gather}
\begin{aligned}
    \widehat{\mu}=\frac{1}{n}\sum_{i=1}^n X_{k,i}&\tanh\frac{X_{k,i}}{\alpha}, \quad \widehat{\sigma}^2=\alpha\widehat{\mu},\\
    \widehat{\mu}^2+\widehat{\sigma}^2&=\frac{1}{n}\sum_{i=1}^n X_{k,i}^2.
\end{aligned}    \label{eq:gmmjointml}
\end{gather}
As defined in Lemma \ref{lem:jointmlsolution}, let the joint ML estimates obtained by solving the above equations be denoted $\widehat{\mu}_{ML}$ and $\widehat{\sigma}_{ML}$. Given $M_{k-1}=\mu$, $\Sigma_{k-1}=\sigma$, we have that
$M_{k}=\widehat{\mu}_{ML}(X_{k,1},\ldots,X_{k,n})$, $\Sigma_{k}=\widehat{\sigma}_{ML}(X_{k,1},\ldots,X_{k,n})$
satisfy the conditions in \eqref{eq:gmmjointml}. So, given $M_k=\mu$, $\Sigma_k=\sigma$, we see that
\begin{equation*}
    M_{k}^2+\Sigma_{k}^2=\frac{1}{n}\sum_{i=1}^n X_{k,i}^2.
\end{equation*}
Since $\EE[X_{k,i}^2\,\vert\,M_k=\mu,\Sigma_k=\sigma]=\mu^2+\sigma^2$, it is clear that $\EE[M_{k+1}^2+\Sigma_{k+1}^2\,\vert\,M_k,\Sigma_k]=M_k^2+\Sigma_k^2$, which implies that $\{M_k^2+\Sigma_k^2\}$ is a martingale under the filtration generated by $\{M_k,\Sigma_k\}$. Since $M_k^2+\Sigma_k^2$ is non-negative and bounded in expectation, we have that $M_k^2+\Sigma_k^2$ converges almost surely to a finite random variable meaning that every sample path of $\{M_k^2+\Sigma_k^2\}$ converges to a finite value. We see that $\Sigma_k\to0$ and $M_k\to$ $($some constant$)$ are possible points of convergence for the sample paths, which suggests model collapse in GMM recursive training. 

To study model collapse, we need to upper bound the variance estimate $\widehat{\sigma}^2_{ML}$. Given the unwieldy $\tanh$ function in the equations, the best upper bound we have is $\widehat{\sigma}^2_{ML}\le \widehat{\sigma}^2_\infty$, which is described in Lemma \ref{lem:jointmlsolution}. To proceed further, we consider a well-motivated approximation of the joint ML estimator where
\begin{equation*}
    \widehat{\mu}\approx \widehat{\mu}_{a} \triangleq \frac{a\widehat{\mu}_0^2}{\sqrt{a^2\widehat{\mu}_0^2 + \left(\dfrac{a\widehat{\mu}_0^2}{\widehat{\sigma}^2_\infty} + 1\right)^2\alpha^2} - \alpha},
\end{equation*}
where $a\ge2$ is a parameter, $\widehat{\mu}_0=\frac{1}{n}\sum_{i=1}^n|X_i|$ and $\widehat{\sigma}^2_\infty=\frac{1}{n}\sum_{i=1}^n X^2_i$. More details on the approximation are in Appendix~\ref{sec:approxjointml}. The error in approximation is observed to be very minimal in simulations (Appendix~\ref{sec:gmmcomparison}). The approximate joint ML estimates with parameter $a$, denoted $\widehat{\mu}_{a,ML}$ and $\widehat{\sigma}_{a,ML}$, can be explicitly solved and characterised as shown in Lemma \ref{lem:jointmlapprox}. Letting 
$S_k^2=\frac{1}{n}\sum_{i=1}^n X_{k,i}^2-\left(\frac{1}{n}\sum_{i=1}^n |X_{k,i}|\right)^2$ 
be the variance of the absolute values of the GMM samples in the $k$-th iteration, Lemma \ref{lem:jointmlapprox} shows that the variance estimate of the approximate joint ML estimator $\widehat{\sigma}^2_{a,ML}(X_{k,1},\ldots,X_{k,n})$ satisfies
\begin{equation}\label{eq:gmmvarbnd}
    S_k^2\le \widehat{\sigma}^2_{a,ML}(X_{k,1},\ldots,X_{k,n})\le (1+\kappa)S_k^2,
\end{equation}
for suitable values of $a$ and $\kappa\in(0,1)$.

\begin{figure}[t]
    \centering
       \conf{\resizebox{0.9\linewidth}{!}}
     \arxiv{\resizebox{0.5\linewidth}{!}}{\input{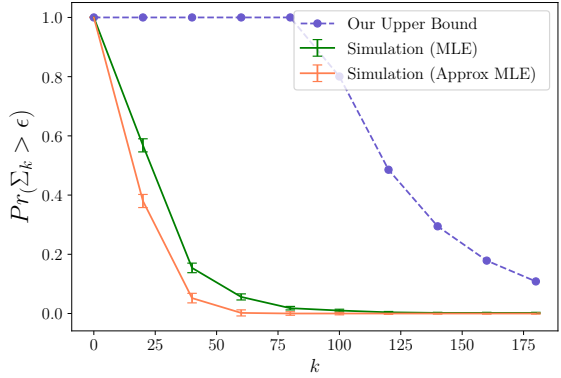}}
    \caption{Comparison of theoretical upper bound with empirical observations of  $\text{Pr}(\Sigma_k>\epsilon)$ in GMM recursive training process $(\mu_0 = 1, \sigma_0 = 1, \epsilon = 10^{-1}, n = 10)$ with Joint ML and Approximate Joint ML estimators.}
    \label{fig:gmmprobbounds}
\end{figure}

Interestingly, both the joint ML estimator and its approximate version are algebraically identical if $X_{k,i}$ is replaced by $|X_{k,i}|$. For $X_{k,i}\sim \frac{1}{2}N(-\mu,\sigma^2)+\frac{1}{2}N(\mu,\sigma^2)$, we have that $|X_{k,i}|\sim |N(\mu,\sigma^2)|$. So, for the purpose of variance analysis, we can assume that $X_{k,1},\ldots,X_{k,n}\sim$ iid $N(\mu,\sigma^2)$. Since $\frac{1}{n}\sum_{i=1}^n |X_{k,i}|\ge\overline{X}_k\triangleq\frac{1}{n}\sum_{i=1}^n X_{k,i}$, we have
$$S^2_k\le \frac{\sum\limits_{i=1}^n X_{k,i}^2-\overline{X}_k^2}{n}\sim \sigma^2\frac{\chi^2_{n-1}}{n-1}.$$
Using the above in \eqref{eq:gmmvarbnd}, we have that
$$
\Sigma_{k+1}^2=\widehat{\sigma}^2_{a,ML}(X_{k,1},\ldots,X_{k,n}) \le (1+\kappa)\Sigma^2_k\,\frac{\chi^2_{n-1}}{n-1}.
$$
Using the above bound on the variance of the approximate ML estimator, we prove the following theorem.
\begin{theorem}
    For the GMM recursive training process defined above with the approximate joint ML estimator given in Algorithm \ref{alg:approx} of Appendix \ref{sec:gmmthmproof} and $n\ge3$,
    \begin{equation*}
        \text{Pr}(\Sigma_k>\epsilon)\le \frac{\sigma_0}{\epsilon}\exp\left\{\frac{-k}{4n}\right\}.
    \end{equation*}
    This implies that the parameter $\sigma$ in the GMM collapses to $0$ almost surely under recursive training.
    \label{thm:gmm}
\end{theorem}
A proof can be found in Appendix \ref{sec:gmmthmproof}. We have $\text{Pr}(\Sigma_k>\epsilon)<\delta$ if 
$$k>4n\log \frac{\sigma_0}{\epsilon\delta},$$
which grows linearly in $n$.  In Figure~\ref{fig:gmmprobbounds}, we show $\text{Pr}(\Sigma_k>\epsilon)$ versus $k$ for the joint ML estimator and the approximate ML estimator and the upper bound from the above Theorem. We report additional experimental results in Appendix~\ref{app:experiments}.

\section{Discussion}
\label{sec:conclusion}

In this paper, we studied the problem of recursive training, where at each round, models are trained based on samples generated by the model in the previous round. We theoretically characterized the time to collapse for a range of fundamental distributions and provide experimental evaluations. Bridging the gap between our upper bound and the exact time to collapse for the Gaussian distribution and analyzing the time to model collapse for the maximum likelihood estimator for the Gaussian mixtures would be interesting future directions. Another potential future direction is to extend our analysis to scenarios where, at each round, models are trained based on a combination of synthetic and human-generated data.
\bibliography{refs}
\bibliographystyle{plainnat}


\appendix
\section{Related work}
\label{sec:app_related}

The recursive training iteration in \eqref{eq:rectrain2} can be termed a stochastic recursion, which is a broad topic with applications ranging from dynamical systems and optimization to reinforcement learning. One type of stochastic recursion, popularly known as stochastic approximation \citep{borkar2022stochastic,Kushner2010}, is well-studied and considers iterations of the form $\theta_{k+1}=\theta_k+a_k(f(\theta_k)+\zeta_k)$, where $\zeta_k$ is random noise or measurement error. 
Another well-studied type of stochastic recursion is known as random function iteration or iterated function system \citep{Diaconis99, Stenflo2012}, which considers an iteration of the form $X_{k+1}=F_{\phi_k}(X_k)$, where $\phi_k$ is an iid sequence that randomly indexes a set of possible functions $F_{\phi_k}$ in each iteration. 

\cite{hataya2023will, martinez2023combining, martinez2023towards} studied the problem of recursive training, where at each round, training data comprises of both human generated data and synthetic data. They showed that increasing the amount of synthetic data reduces the model quality over time in image generation tasks.  \cite{shumailov2023curse, shumailov2024ai} investigated recursive training with only synthetic data in each round and empirically demonstrated model collapse on variational autoencoders, Gaussian mixtures, and language models. They further provided theoretical intuition for model collapse. \cite{alemohammad2023self} empirically analyzed recursive training with fully synthetic data and also with combinations of synthetic and human generated data. \cite{bertrand2023stability} 
theoretically analyzed the effect of training generative models on a combination of real and synthetic datasets and validated their results empirically.  \cite{gerstgrasser2024model} demonstrated empirically that combining synthetic data with original data avoids model collapse across several deep generative models. \cite{zhang2024regurgitative} proposed several ways to reduce performance loss when training with combination of synthetic and human generated data for language models. \cite{dohmatob2024model} theoretically analyzed model collapse for high-dimensional regression. \cite{jain2024scaling, dohmatob2024tale} examined how scaling laws change when synthetic data is taken into account. \cite{marchi2024heat} studied how that temperature dependent sampling affects recursive training.  \cite{seddik2024bad} theoretically proved that model collapse cannot be avoided when solely on synthetic data.

\section{Analysis for Poisson and discrete distributions}

We first prove auxiliary results that would be helpful in proving our main theorems. 

\begin{lemma}
For any positive integer $n$ and $a \in[0, 1]$,
\begin{equation*}
  e^{-an}(1-na^2) \leq (1-a)^n \leq e^{-an}.
\end{equation*}
\label{lem:a_bound}
\end{lemma}
\begin{proof}
When $a = 0$, $1 - a = e^{-a} = 1$ and the derivative of $1
- a$ is never larger than the derivative of $e^{-a}$ for all $a \geq 0$. Hence for any positive integer $n$ and $a \geq 0$, $(1-a)^n \leq e^{-an}$. Similarly, it can be shown that $e^{an} \geq
(1+a)^n$. Hence $(1-a)^n e^{an} \geq (1-a)^n (1+a)^n = (1-a^2)^n \geq 1 - na^2$, thus completing the proof.
\end{proof}
\begin{lemma}
\label{lem:poisson1}
Let $g(x)=1-e^{-x}$ and $g^{\circ i}$ denotes $g$ composed $i$ times. Then, for $i \geq 1$,
\[
\frac{1}{i} \leq g^{\circ i}(\infty) \leq \frac{3}{i}.
\]
\end{lemma}
\begin{proof}
    It is easy to see that $g(\infty) = 1$. For $i \geq 2$, we show by induction. 
\begin{align*}
g^{\circ i}(\infty) & = 1 - e^{-g^{i-1}(\infty)}  \\
& \leq 1 - e^{-3/(i-1)}  \\
& \leq \frac{3}{(i-1)} - \frac{9}{3 \cdot (i-1)^2}  \\
& = \frac{3}{(i-1)} - \frac{3}{(i-1)^2}  \\
& = \frac{3}{(i-1)} - \frac{3}{i(i-1)}  \\
& = \frac{3}{i},
\end{align*}
where the last inequality uses the fact that $1-e^{-x} \leq x - x^2/3$, for $x\leq 1$.

For the other direction, observe that $i=1$ the result holds as $g(\infty) = 1$. For $i \geq 2$, we show by induction,
\begin{align*}
g^{\circ i}(\infty) & = 1 - e^{-g^{i-1}(\infty)}  \\
& \geq 1 - e^{-1/(i-1)}  \\
& \geq \frac{1}{(i-1)} - \frac{1}{2 \cdot (i-1)^2}  \\
& \geq \frac{1}{(i-1)} - \frac{1}{i(i-1)}  \\
& = \frac{1}{i},
\end{align*}
where the penultimate inequality uses the fact that $1-e^{-x} \geq x - x^2/2$, for $x\leq 1$.
\end{proof}
We now have all the tools to prove the main result for the main result for Poisson distributions.
\begin{proof}[Proof of Theorem~\ref{thm:poisson}]
Recall that $\tilde{\Lambda}_k=n\Lambda_k$ and we have the recursion: $\tilde{\Lambda}_{k+1}\sim \text{Poisson}(\tilde{\Lambda}_k)$.
    Hence the convergence is possible only to $0$ and, so $ \tilde{\Lambda}_{k}\to0$ almost surely as $k\to\infty$. To establish a rate bound, we compute and bound $\Pr( \tilde{\Lambda}_{i}=0)$ for an arbitrary $i$ as follows. 
\begin{align}
    \Pr(\tilde{\Lambda}_i=0)&=\sum_{k=0}^{\infty}\Pr(\tilde{\Lambda}_{i-1}=k)\Pr(\tilde{\Lambda}_i=0|\tilde{\Lambda}_{i-1}=k)\nonumber\\
    &=\sum_{k=0}^{\infty}\Pr(\tilde{\Lambda}_{i-1}=k)e^{-k}\label{eq:posson1}\\
    &=\sum_{k=0}^{\infty}\left(\sum_{l=0}^{\infty}\Pr(\tilde{\Lambda}_{i-2}=l)\Pr(\tilde{\Lambda}_{i-1}=k|\tilde{\Lambda}_{i-2}=l)\right)e^{-k}\nonumber\\
    &=\sum_{l=0}^{\infty}\Pr(\tilde{\Lambda}_{i-2}=l)e^{-l}\left(\sum_{k=0}^{\infty}\frac{(le^{-1})^k}{k!}\right)\nonumber\\
    &=\sum_{l=0}^{\infty}\Pr(\tilde{\Lambda}_{i-2}=l)\exp(-l(1-e^{-1})).\label{eq:poisson2}
\end{align}
From the forms of \eqref{eq:posson1} and \eqref{eq:poisson2}, we assume the induction hypothesis that
\begin{equation}
\label{eq:poissonind}
\Pr(\tilde{\Lambda}_i=0)=\sum_{k=0}^{\infty}\Pr(\tilde{\Lambda}_{i-j}=k)\exp(-kg_j),
\end{equation}
where $f(x)=1-e^{-x}$ and $g^{\circ j}$ is $f$ composed $j$ times, for $j=1,2,\ldots$. The base case of $k=1,2$ have been proved. So, we assume that \eqref{eq:poissonind} is true for an arbitrary $j$ and proceed to prove for $j+1$ as follows.
\begin{align}
    \Pr(\tilde{\Lambda}_i=0)&=\sum_{k=0}^{\infty}\left(\sum_{l=0}^{\infty}\Pr(\tilde{\Lambda}_{i-j-1}=l)\Pr(\tilde{\Lambda}_{i-j}=k|\tilde{\Lambda}_{i-j-1}=l)\right)\exp(-kg_j)\nonumber\\
    &=\sum_{l=0}^{\infty}\Pr(\tilde{\Lambda}_{i-j-1}=l)e^{-l}\sum_{k=0}^{\infty}\exp(-kg_j)\frac{l^k}{k!}\nonumber\\
    &=\sum_{l=0}^{\infty}\Pr(\tilde{\Lambda}_{i-j-1}=l)e^{-l}\sum_{k=0}^{\infty}\frac{(l\exp(-g_j))^k}{k!}\nonumber\\
    &=\sum_{l=0}^{\infty}\Pr(\tilde{\Lambda}_{i-j-1}=l)e^{-l}\exp(l\exp(-g_j))\nonumber\\
    &=\sum_{l=0}^{\infty}\Pr(\tilde{\Lambda}_{i-j-1}=l)\exp(-l(1-\exp(-g_j)))\nonumber\\
    &=\sum_{l=0}^{\infty}\Pr(\tilde{\Lambda}_{i-j-1}=l)\exp(-lg_{j+1}).\nonumber
\end{align}
This completes the induction step of the proof.

Since $\tilde{\Lambda}_0=n \lambda$, using $j=i$ in \eqref{eq:poissonind} results in 
\begin{equation}
\label{eq:poissonfinal}
\Pr(\tilde{\Lambda}_i=0)=\exp(-n \lambda g_i(\infty)).    
\end{equation}

Using Lemma \ref{lem:poisson1} in \eqref{eq:poissonfinal},
$$\exp(-n\lambda/i) \ge \Pr(\tilde{\Lambda}_i=0)\ge \exp(-3n\lambda/i).$$
This concludes the proof of convergence.
\end{proof}
Before we prove the Bernoulli result, we prove few claims, which would be helpful in proving the result. 
\begin{lemma}
  \label{lem:upper_lambda}
  For any $j \geq 1$ and $\lambda \geq 0$,
  \begin{align*}
  \sum_{i=0}^{n} \binom{n}{i} \left(\frac{\lambda}{n} \right)^i \left( 1- \frac{\lambda}{n} \right)^{n-i} \exp(-ig_{j})
  \leq \exp(-\lambda g_{j+1}),
  \end{align*}
  where $g_j$ is defined in~\eqref{eq:gk}.
\end{lemma}
\begin{proof}
  The proof follows by~Lemma~\ref{lem:a_bound} and the definition of $g$,
\begin{align}
  \sum_{i=0}^{n} \binom{n}{i} \left(\frac{\lambda}{n} \right)^i \left( 1- \frac{\lambda}{n} \right)^{n-i} \exp(-ig_{j}) 
  & =  \left(1 - \frac{\lambda}{n} \left(1 - \frac{1}{e^{g_{j}}} \right) \right)^n \nonumber \\
   & =  \left(1 - \frac{\lambda g_{j+1}}{n} \right)^n \nonumber\\
    & \leq \exp(-\lambda g_{j+1}),\nonumber
\end{align}
where the last inequality uses the fact that $1 - x \leq e^{-x}$ for all $x$.
  \end{proof}
  \begin{lemma}
  \label{lem:lower_lambda}
  For any $j \geq 1$, $c_1 \geq 1$, $c_2 \geq 0$,and $n \geq \lambda \geq 0$,
 \begin{align*}
   & \sum_{i=0}^{n} \binom{n}{i} \left(\frac{\lambda}{n} \right)^i \left( 1- \frac{\lambda }{n} \right)^{n-i} \exp(-ig_j)
\left( 1 - \frac{ c_1i^2 + c_2 i }{n} \left( \sum^j_{m=1} g^2_m \right) 
\right) \\
& \geq \exp(-\lambda g_{j+1}) \left( 1 - \frac{c_1\lambda ^2 + (c_1 + c_2)\lambda}{n} \left( \sum^{j+1}_{m=1} g^2_m \right) \right),
  \end{align*}
    where $g_j$ is defined in~\eqref{eq:gk}.
\end{lemma}
\begin{proof}
\begin{align*}
  &  \sum_{i=0}^{n} \binom{n}{i} \left(\frac{\lambda}{n} \right)^i \left( 1- \frac{\lambda }{n} \right)^{n-i} \exp(-ig_j)
\left( 1 - \frac{c_1i^2 + c_2 i}{n} \left( \sum^j_{m=1} g^2_m \right) 
\right)\nonumber\\
& =  \left( 1 - \frac{\lambda}{n} \left( 1 - \frac{1}{e^{g_j}} \right)\right)^n \EE_{i \sim  \frac{\lambda}{en(1-\lambda(1-1/e^{g_j})/n)}} \left[\left( 1 - \frac{c_1i^2 + c_2 i}{n} \left( \sum^j_{m=1} g^2_m \right) \right)\right] \nonumber \\
  & \stackrel{(a)}{\geq} \left(1 - \frac{\lambda}{n} \left(1 - \frac{1}{e^{g_j}} \right) \right)^n
  \left( 1 - \frac{c_1\lambda^2+ (c_1+c_2) \lambda }{n} \left( \sum^j_{m=1} g^2_m \right) \right) \\
   & =   \left(1 - \frac{\lambda g_{j+1}}{n} \right)^n
\left( 1 - \frac{c_1\lambda^2+ (c_1+c_2) \lambda}{n} \left( \sum^j_{m=1} g^2_m \right) \right) 
  \\
  & \stackrel{(b)}{\geq}  \exp \left(- \lambda g_{j+1} \right)
\left( 1 - \frac{c_1\lambda^2+ (c_1+c_2) \lambda}{n} \left( \sum^j_{m=1} g^2_m \right) \right) \left( 1 - \frac{\lambda^2}{n} g^2_{j+1}  \right)
  \\
    & \stackrel{c)}{\geq} \exp(-\lambda g_{j+1}) \left( 1 - \frac{c_1\lambda^2+ (c_1+c_2) \lambda}{n} \left( \sum^{j+1}_{m=1} g^2_m \right) \right).\nonumber
\end{align*}
 $(a)$ uses the fact that $\EE_{Y \sim \bin(n, p)}[Y^2] = n^2 p^2 + np
(1-p) \leq n^2p^2 + np $ and the fact that $
\frac{\lambda}{en(1-\lambda(1-1/e^{g_j})/n)} \leq \lambda/n$ for all
$\lambda \geq 0$. 
Lemma~\ref{lem:a_bound} implies $(b)$.  $(1-c)(1-d) \geq 1 - c - d$ for all
$c \geq 0$ and $d\geq 0$ and hence $(c)$.
 \end{proof}
We now have all the tools to prove our result for Bernoulli distributions.
\begin{proof}[Proof of Theorem~\ref{th:discrete_main}]
Let $\tilde{P}_k = n P_k$. We compute $\Pr(\tilde{P}_k = 0)$ for
arbitrary $k$ as follows:
\begin{align}
    \Pr(\tilde{P}_k=0)&=\sum_{i=0}^{n}\Pr(\tilde{P}_{k-1}=i)\Pr(\tilde{P}_k=0|\tilde{P}_{k-1}=i)\nonumber\\
    &=\sum_{i=0}^{n}\Pr(\tilde{P}_{k-1}=i) \left( 1 - \frac{i}{n} \right)^n \label{eq:bernoulli1}\\
    &= \sum_{i=0}^{n}\left(\sum_{j=0}^{n}\Pr(\tilde{P}_{k-2}=j)\Pr(\tilde{P}_{k-1}=i|\tilde{P}_{j-2}=j)\right) \left( 1 - \frac{i}{n} \right)^n \nonumber\\
    &= \sum_{i=0}^{n}\left(\sum_{j=0}^{n}\Pr(\tilde{P}_{k-2}=j)\binom{n}{i} \left( \frac{j}{n}\right)^i
    \left(1 - \frac{j}{n}\right)^{n-i}\right) \left( 1 - \frac{i}{n} \right)^n \nonumber\\
    &= \sum_{j=0}^{n}\sum_{i=0}^{n}\Pr(\tilde{P}_{k-2}=j)\binom{n}{i} \left( \frac{j}{n}\right)^i
    \left(1 - \frac{j}{n}\right)^{n-i} \left( 1 - \frac{i}{n} \right)^n \label{eq:bernoulli2}
\end{align}
By~\eqref{eq:bernoulli1} and~Lemma~\ref{lem:a_bound},
\begin{align*}
\sum_{i=0}^n \Pr(\tilde{P}_{k-1}= i) e^{-i} \left( 1- \frac{i^2}{n} \right) \leq \Pr(\tilde{P}_k = 0) \leq \sum_{i=0}^n \Pr(\tilde{P}_{k-1}=i) e^{-i}. 
\end{align*}
Similarly by~\eqref{eq:bernoulli2} and~Lemma~\ref{lem:a_bound},
\begin{align}
  \Pr(\tilde{P}_k = 0)
    &= \sum_{j=0}^{n}\sum_{i=0}^{n}\Pr(\tilde{P}_{k-2}=j)\binom{n}{i} \left( \frac{j}{n}\right)^i
  \left(1 - \frac{j}{n}\right)^{n-i} \left( 1 - \frac{i}{n} \right)^n \nonumber \\
  & \leq  \sum_{j=0}^{n}\sum_{i=0}^{n}\Pr(\tilde{P}_{k-2}=j)\binom{n}{i} \left( \frac{j}{n}\right)^i
  \left(1 - \frac{j}{n}\right)^{n-i} e^{-i} \nonumber \\
  & = \sum_{j=0}^n \Pr(\tilde{P}_{k-2} =j) \left( 1 - \frac{j}{n} \left( 1- \frac{1}{e}\right)\right)^n  \nonumber \\
  & \leq \sum_{j=0}^n \Pr(\tilde{P}_{k-2} = j) e^{-j(1-1/e)} \label{eq:temp_bber} 
\end{align}
We provide another lower bound: 
\begin{align}
  \Pr(\tilde{P}_k = 0)
    &= \sum_{j=0}^{n}\sum_{i=0}^{n}\Pr(\tilde{P}_{k-2}=j)\binom{n}{i} \left( \frac{j}{n}\right)^i
  \left(1 - \frac{j}{n}\right)^{n-i} \left( 1 - \frac{i}{n} \right)^n \nonumber \\
     &= \sum_{j=0}^{n}\Pr(\tilde{P}_{k-2}=j) \sum_{i=0}^{n}\binom{n}{i} \left( \frac{j}{n}\right)^i
  \left(1 - \frac{j}{n}\right)^{n-i} \left( 1 - \frac{i}{n} \right)^n \nonumber \\
     &\stackrel{(a)}{\ge} \sum_{j=0}^{n}\Pr(\tilde{P}_{k-2}=j) \sum_{i=0}^{n}\binom{n}{i} \left( \frac{j}{n}\right)^i
  \left(1 - \frac{j}{n}\right)^{n-i} e^{-i} \left( 1 - \frac{i^2}{n} \right) \nonumber \\
  & \stackrel{(b)}{\geq} \sum^n_{j=0} \Pr(\tilde{P}_{k-2}=j) e^{-j(1-1/e)} \left( 1 - \frac{j^2 + j}{n} \left( 1 + \left(1 - \frac{1}{e}\right)^2 \right) \right), \label{eq:temp_cber}
\end{align}
Here $(a)$ follows from~Lemma~\ref{lem:a_bound} and $(b)$ follows from Lemma~\ref{lem:lower_lambda}.

From the forms of \eqref{eq:temp_bber} and \eqref{eq:temp_cber}, we assume the induction hypothesis that
\begin{align}
& \sum_{i=0}^{n}\Pr(\tilde{P}_{k-j}=i)\exp(-ig_j) \left( 1 - \frac{i^2 + j i}{n} \left( \sum^j_{m=1}g^2_m \right) \right) \nonumber \\
& \leq \Pr(\tilde{P}_k=0) \leq \sum_{i=0}^{n}\Pr(\tilde{P}_{k-j}=i)\exp(-ig_j), 
\label{eq:binomind} 
\end{align}
The base case of $j=1,2$ have been proved. So, we
assume that \eqref{eq:binomind} is true for an arbitrary $j$ and proceed to prove for $j+1$ as follows. We first focus on the lower bound.
\begin{align}
  &  \Pr(\tilde{P}_k=0)  \nonumber \\
  &\geq\sum_{i=0}^{n}\left(\sum_{l=0}^{n}\Pr(\tilde{P}_{k-j-1}=l)\Pr(\tilde{P}_{k-j}=i|\tilde{P}_{i-j-1}=l)\right)\exp(-ig_j)
\left( 1 - \frac{i^2 + j i}{n} \left( \sum^j_{m=1}g^2_m \right) \right)
  \nonumber\\
  & =  \sum_{i=0}^{n}\left(\sum_{l=0}^{n}\Pr(\tilde{P}_{k-j-1}=l) \binom{n}{i} \left(\frac{l}{n} \right)^i \left( 1- \frac{l}{n} \right)^{n-i} \right)\exp(-ig_j)
\left( 1 - \frac{i^2 + j i}{n} \left( \sum^j_{m=1}g^2_m \right) \right)
  \nonumber\\
  & =  \sum_{l=0}^{n}\Pr(\tilde{P}_{k-j-1} = l) \left(\sum_{i=0}^{n} \binom{n}{i} \left(\frac{l}{n} \right)^i \left( 1- \frac{l}{n} \right)^{n-i} \exp(-ig_j)
\left( 1 - \frac{i^2 + j i}{n} \left( \sum^j_{m=1}g^2_m \right) \right)
  \right)\nonumber\\
  & \stackrel{(a)}{\geq} \sum_{l=0}^{n}\Pr(\tilde{P}_{k-j-1}=l)\exp(-lg_{j+1}) \left( 1 - \frac{l^2 + (j+1)l}{n} \left( \sum^{j+1}_{m=1}g^2_m \right) \right),\nonumber
\end{align}
where $(a)$ follows from Lemma~\ref{lem:lower_lambda}.  We now prove
the bound in the other direction, which follows from Lemma~\ref{lem:upper_lambda}.
\begin{align}
  \Pr(\tilde{P}_k=0) \nonumber & \leq \sum_{i=0}^{n}\left(\sum_{l=0}^{n}\Pr(\tilde{P}_{k-j-1}=l)\Pr(\tilde{P}_{k-j}=i|\tilde{P}_{i-j-1}=l)\right)\exp(-ig_j)\nonumber\\
  & =  \sum_{i=0}^{n}\left(\sum_{l=0}^{n}\Pr(\tilde{P}_{k-j-1}=l) \binom{n}{i} \left(\frac{l}{n} \right)^i \left( 1- \frac{l}{n} \right)^{n-i} \right)\exp(-ig_j)\nonumber\\
  & =  \sum_{l=0}^{n}\Pr(\tilde{P}_{k-j-1} = l) \left(\sum_{i=0}^{n} \binom{n}{i} \left(\frac{l}{n} \right)^i \left( 1- \frac{l}{n} \right)^{n-i} \exp(-ig_j) \right)\nonumber\\
    & \leq \sum_{l=0}^{n}\Pr(\tilde{P}_{k-j-1}=l)\exp(-lg_{j+1}).\nonumber
\end{align}
This completes the induction step of the proof. Using $\tilde{P}_0=n p_0$, and using $j=k$ in \eqref{eq:binomind} results in 
\begin{align*}
  \exp(- n p_0 g_k )\left(1 - \frac{(np_0)^2 +  n p_0 k}{n} \left( \sum^k_{m=1}g^2_m \right) \right) \leq \Pr(\tilde{P}_k=0) \leq \exp(- n p_0  g_k)
\end{align*}
The proof follows by numerically showing that $\sum^k_{m=1}g^2_m \leq 3 $.
\end{proof}

\subsection{Absorption probabilities for Bernoulli model}
\label{sec:bernoulliabsorption}
We will follow the standard method for calculation of absorption probabilities in Markov chains. The Markov chain $\{nP_k\}$, $k=1,2,\ldots$, has states $\{0,1,\ldots,n\}$ with transition probabilities 
\begin{align*}
\text{Pr}(0\to j)=\begin{cases}
    1&j=0\\
    0&j\ne0
\end{cases},\qquad \text{Pr}(n\to j)=\begin{cases}
    1&j=n\\
    0&j\ne n
\end{cases},\\
\text{Pr}(i\to j)=\binom{n}{j}(i/n)^j(1-i/n)^{n-j},\ i\notin \{0, n\}.
\end{align*}
Let $a_i=\text{Pr}(\text{absorption to }0|nP_1=i)$ be the conditional probability of the chain eventually getting absorbed to $0$ given that $nP_1=i$. For each $i$, we have
\begin{align*}
a_i&=\sum_{j=0}^n \text{Pr}(\text{absorption to }0,nP_2=j|nP_1=i)\\
&=\sum_{j=0}^n \text{Pr}(nP_2=j|nP_1=i)\text{Pr}(\text{absorption to }0|nP_2=j)\\
&=\sum_{j=0}^{n-1} \binom{n}{j}(i/n)^j(1-i/n)^{n-j}a_j.
\end{align*}
Substituting $a_j=(n-j)/n$, we get
\begin{align*}
    a_i&=\sum_{j=0}^{n-1} \binom{n}{j}\frac{n-j}{n}(i/n)^j(1-i/n)^{n-j}\\
    &=(1-i/n)\sum_{j=0}^{n-1} \binom{n-1}{j}(i/n)^j(1-i/n)^{n-1-j}\\
    &=(n-i)/n\quad\text{for all }i.
\end{align*}
Now, $P_0=p_0$ and Pr$(nP_1=i)=\binom{n}{i}p_0^i(1-p_0)^{n-i}$. So,
$$\text{Pr}(\text{absorption to }0)=\sum_{i=0}^n a_i\binom{n}{i}p_0^i(1-p_0)^{n-i}=1-p_0$$
by the same process as the previous summation for $a_i$. A very similar calculation shows that the probability of absorption to the State $n$ is $p_0$.

\subsection{Additional experiments for Bernoulli models}
\label{sec:seddick}

In this section, we compare our results for Bernoulli models to that of \cite{seddik2024bad}. Recall that \cite{seddik2024bad} showed that 
$\Pr(P_k \notin \{0, 1\})$ lies in 
\begin{align*}
4p_0(1-p_0) \left( 1 - \frac{1}{n}\right)^k \leq \Pr(P_k \notin \{0, 1\}) \leq   2np_0(1-p_0)  \left( 1 - \frac{1}{n}\right)^k.
\end{align*}
By Theorem~\ref{th:discrete_main}, 
\begin{align*}
   \max\left(1 - 3np^2_0 - 3 k p_0, 0 \right) \exp(- n p_0 g_k ) 
   & \leq   \text{Pr}(P_k = 0) \leq \exp(-n p_0 g_k), \\
    \max\left(1 - 3n(1-p_0)^2 - 3 k (1-p_0), 0 \right) \exp(- n (1-p_0) g_k ) 
   & \leq   \text{Pr}(P_k = 1)  \leq \exp(-n (1-p_0) g_k).
\end{align*}
Combining the above two equations, we get that
\begin{align*}
1 - \left( \exp(-n (p_0) g_k) +  \exp(- n (1-p_0) g_k ) \right) \leq \Pr(P_k \notin \{0, 1\}),
\end{align*}
and
\begin{align*}
1 - \left(   \max\left(1 - 3np^2_0 - 3 k p_0, 0 \right) e^{- n p_0 g_k }  +   \max\left(1 - 3n(1-p_0)^2 - 3 k (1-p_0), 0 \right) e^{- n (1-p_0) g_k }\right) \geq \Pr(P_k \notin \{0, 1\}).
\end{align*}

\begin{figure}[ht]
    \centering
     \resizebox{0.75\linewidth}{!}{\input{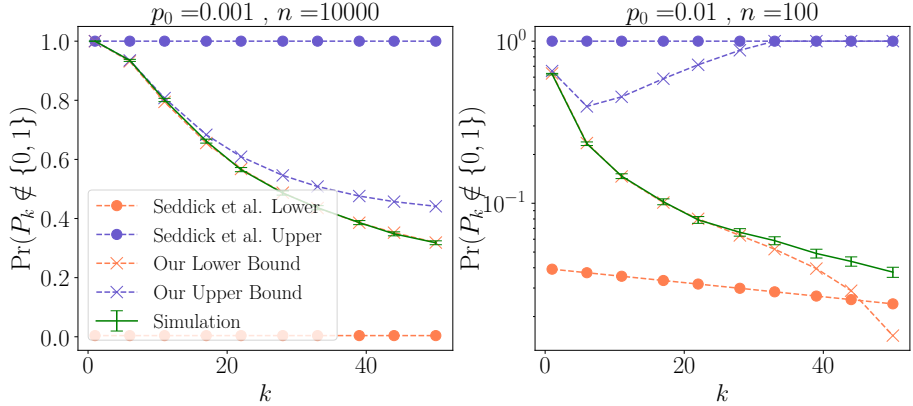}}
    \caption{Comparison of theoretical bounds and empirical observations for $\Pr(P_k \notin \{0,1\})$ at a particular generation in a Bernoulli recursive training process with $n$ samples at each generation. Results are shown for varying initial success probabilities $(p_0)$.}
    \label{fig:bernoulli_with_seddick}
\end{figure}

We compare our results with empirical observations in Figure \ref{fig:bernoulli_with_seddick} for different values of $p_0,n$.  Observe that for most values of $k$, our lower bound is the tightest of all the bounds.

\subsection{Discrete distribution under Poisson sampling}
\label{sec:poisson_discrete}
Consider a discrete distribution on the alphabet $\{1,2,\ldots,m\}$ with $p_i$ being the probability of $i$. Under Poisson$(n)$ sampling, the number of $i$'s generated is distributed as Poisson$(np_i)$ independently for each $i$. 

Based on the above, recursive Poisson sampling of a discrete distribution may be modelled as follows. The parameter vector is $n=(n_1,\ldots,n_m)$, where $n_i\in\{0,1,2,\ldots\}$, is a non-negative integer. The sampling distribution $P_n$ is iid Poisson$(n_i)$, $i=1,\ldots,m$. The recursive training process $\{N_k\}$ is initialized as $N_0=(n_{0,1},\ldots,n_{0,m})$. Given $N_k=(n_1,\ldots,n_m)$, the samples in the $k$-th round $X_{k,i}$ are iid Poisson$(n_i)$, $i=1,\ldots,m$. The ML estimate of $n_i$ is given as $\widehat{n}_i=X_{k,i}\sim$ Poisson$(n_i)$. So, we obtain the recursion
$$N_{k+1}\sim\text{ Poisson}(N_k).$$
So, the above recursive Poisson sampling process of a discrete distribution is exactly the Poisson recursive training process considered in Section \ref{sec:poisson}. So, the same model collapse results that were derived in Section \ref{sec:poisson} apply for the recursive Poisson sampling of a discrete distribution. Hence we can bound number of distinct symbols after $k$ steps as follows.
\begin{corollary}
Let $\text{uniq}_k= \sum^m_{i=1}I(\widehat{\Theta}_{k,i} \neq 0)$ denote that number of distinct elements after $k$ steps in the distribution distribution estimation under Poisson sampling process. Then
\[
\EE\left[\text{uniq}_k \right] = \sum^m_{i=1} \left(1 - \exp(n \cdot \theta_{0, i} \cdot g_k \right)),
\]
where $g_k$ is defined in~\eqref{eq:gk}.
\end{corollary}

\section{Gaussian recursive training: Proof of Theorem \ref{thm:Gaussian}}
\label{sec:Gaussian}
We have $\Sigma^2_0=\sigma^2$ and the recursion 
$$\Sigma^2_k=\Sigma^2_{k-1}U_k=\sigma^2_0U_1U_2\cdots U_k,$$ 
where $U_1,\ldots,U_k\sim\text{ iid }\chi^2_m/m$ (setting $m=n-1$) and the common density of $U\sim\chi^2_m/m$ is given by 
$$f_U(u)=\frac{(m/2)^{m/2}}{\Gamma(m/2)}u^{\frac{m}{2}-1}e^{-mu/2},\ u>0,$$
where $\Gamma(z)\triangleq\int_0^{\infty}x^{z-1}e^{-x}dx$ is the standard Gamma function. A direct calculation shows that 
$$\EE[U^t]=\int_0^\infty\frac{(m/2)^{m/2}}{\Gamma(m/2)}u^{t+\frac{m}{2}-1}e^{-mu/2}du=\frac{\Gamma(t+m/2)}{(m/2)^t\Gamma(m/2)}.$$
By log-convexity property and other standard properties of Gamma functions, we have that, for $m/2\ge1$, 
$$\EE[U^t]:\begin{cases}
=1&t=0,1,\\
<1&0<t<1,\\
>1&t>1.
\end{cases}$$
The area of deriving bounds for ratios of Gamma functions is a rich one \citep{GammaBoundsQi2013}. For $t=1/2$, we use Gurland's bound \citep{GurlandGamma1956} $\Gamma((m+1)/2)/\Gamma(m/2)<m/\sqrt{2m+1}$ and simplify to get
\begin{equation}
    E[U^{1/2}]\le \left(1+\frac{1}{2m}\right)^{-1/2}\overset{(a)}{\le} \exp\left\{-\frac{1}{4m}\left(1-\frac{1}{2m}\right)\right\}\overset{(b)}{\le}\exp\left\{-\frac{1}{4m+3}\right\},
    \label{eq:EsqrtUbound}
\end{equation}
where we use $(1+x)^{-1}\le 1-x+x^2\le e^{-x+x^2}$ for the inequality $(a)$ and $\frac{1}{4m}-\frac{1}{8m^2}>\frac{1}{4m+3}$, $m\ge2$ for $(b)$.

Since $\EE[(\Sigma^2_k)^t]=(\sigma^2_0)^t(E[U^t])^k$, by Markov inequality, we get that 
\begin{equation}
\text{Pr}(\Sigma_k\ge\epsilon)=\text{Pr}((\Sigma^2_k)^t\ge(\epsilon^2)^t)\le \min_{t\in[0,1]}\left(\frac{\sigma^2_0}{\epsilon^2}\right)^t\left(\frac{\Gamma(t+m/2)}{(m/2)^t\Gamma(m/2)}\right)^k\le \frac{\sigma_0}{\epsilon}\exp\left\{-\frac{k}{4m+3}\right\},
\end{equation}
where we set $t=1/2$ and use the bound in \eqref{eq:EsqrtUbound} for the last inequality. This completes the proof of Theorem \ref{thm:Gaussian}.

\section{Analysis of Gaussian mixture models}

\subsection{Joint ML estimation in a Gaussian mixture model (GMM)}
\begin{figure}[htb]
    \centering
   \conf{ \resizebox{\linewidth}{!}}   \arxiv{ \resizebox{0.8\linewidth}{!}}{\input{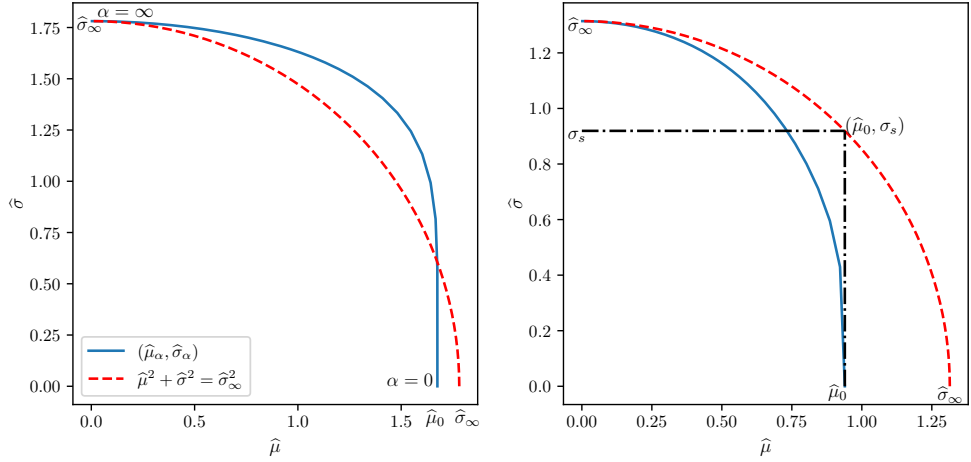}}
    \caption{Illustrative plots of the two curves in Lemma \ref{lem:gmmjointml} for two sets of samples with $n=10$, $\mu=1$, $\sigma=1$.}
    \label{fig:gmmcurves}
\end{figure}

\begin{lemma}
    For $X_1,X_2,\ldots,X_n\sim$ iid $\frac{1}{2}N(-\mu,\sigma^2)+\frac{1}{2}N(\mu,\sigma^2)$, the joint maximum-likelihood estimates of $\mu\ge0$ and $\sigma\ge0$ from $n$ samples $x_1,\ldots,x_n$ in a particular instance are given as the point of intersection of the following two curves in the $\widehat{\mu}-\widehat{\sigma}$ plane:
    \begin{align}
        (\widehat{\mu}_\alpha,\widehat{\sigma}_\alpha):& \ \widehat{\mu}_\alpha \triangleq \frac{1}{n}\sum_{i=1}^n x_i\tanh \frac{x_i}{\alpha},\ \widehat{\sigma}_\alpha^2 \triangleq \alpha \widehat{\mu}_\alpha \text{   for   }\alpha\in(0,\infty),\label{eq:gmmcurve1}\\ 
        \text{circle}:& \ \widehat{\mu}^2+\widehat{\sigma}^2 = \frac{1}{n}\sum_{i=1}^n x_i^2.\label{eq:gmmcircle}
    \end{align}
\label{lem:gmmjointml}
\end{lemma}
\begin{proof}
The likelihood function $L$ is
\begin{align*}
L&=\prod_{i=1}^n\frac{1}{2\sqrt{2\pi}\sigma}\left(e^{-(x_i+\mu)^2/2\sigma^2}+e^{-(x_i-\mu)^2/2\sigma^2}\right)\nonumber\\
&= \frac{1}{(2\sqrt{2\pi})^n} \frac{e^{-n\mu^2/2\sigma^2}}{\sigma^n}e^{-(\sum_{i=1}^n x_i^2)/2\sigma^2}\prod_{i=1}^n\left(e^{\mu x_i/\sigma^2}+e^{-\mu x_i/\sigma^2}\right).
\end{align*}
So, the log-likelihood function is
\begin{align*}
    \log L & = n\log\frac{1}{2\sqrt{2\pi}}-n\log\sigma-n\frac{\mu^2}{2\sigma^2}-\frac{1}{2\sigma^2}\sum_{i=1}^n x_i^2+\sum_{i=1}^n\log\left(e^{\mu x_i/\sigma^2}+e^{-\mu x_i/\sigma^2}\right).
\end{align*}
Differentiating partially with respect to $\mu$ and $\sigma$,
\begin{align*}
    \frac{\partial\log L}{\partial \mu} &= \frac{-n\mu}{\sigma^2}+\frac{1}{\sigma^2}\sum_{i=1}^n x_i\tanh\frac{\mu x_i}{\sigma^2},\\
    \frac{\partial\log L}{\partial \sigma} &= \frac{-n}{\sigma}+\frac{n\mu^2}{\sigma^3}+\frac{1}{\sigma^3}\sum_{i=1}^n x_i^2-\frac{2\mu}{\sigma^3}\sum_{i=1}^n x_i\tanh\frac{\mu x_i}{\sigma^2}.
\end{align*}
Equating the above two to $0$, simplifying and setting the parameter $\alpha=\sigma^2/\mu$ results in the implicit equations given above. 
\end{proof}
For the $n$ GMM samples $x_1,\ldots,x_n$, we define the following sample statistics using the functions $\widehat{\mu}_\alpha$ and $\widehat{\sigma}_\alpha$ defined in the above lemma.
\begin{itemize}
    \item Absolute mean: $\widehat{\mu}_0 = \frac{1}{n}\sum_{i=1}^n|x_i|$
    \item Second moment: $\widehat{\sigma}^2_\infty = \frac{1}{n} \sum_{i=1}^n x_i^2$
    \item Variance of absolute value of samples: $\sigma^2_s=\frac{1}{n}\sum_{i=1}^nx_i^2-\left(\sum_{i=1}^n|x_i|\right)^2=\widehat{\sigma}^2_\infty-\widehat{\mu}_0^2$
\end{itemize}
Figure \ref{fig:gmmcurves} shows the plots of the two curves appearing in Lemma \ref{lem:gmmjointml} for two typical cases.
The blue solid line is the curve $(\widehat{\mu}_\alpha,\widehat{\sigma}_\alpha)$ parameterized by $\alpha\in(0,\infty)$. The red dotted line is the circle $\widehat{\mu}^2+\widehat{\sigma}^2=\widehat{\sigma}^2_\infty$. The corner points of the two curves are shown in the figure. The circle goes from $(\widehat{\sigma}_\infty,0)$ on the $\widehat{\mu}$ axis to $(0,\widehat{\sigma}_\infty)$ on the $\widehat{\sigma}$ axis. The parameterized curve goes from $(\widehat{\mu}_0,0)$ for $\alpha=0$ on the $\widehat{\mu}$ axis to $(0,\widehat{\sigma}_\infty)$ for $\alpha=\infty$ on the $\widehat{\sigma}$ axis. The point $(\widehat{\mu}_0,\sigma_s)$ is on the circle and is shown in the plot to the right.

In the plot to the left, the two curves intersect at a point where $\alpha$ is finite and also at $(0,\widehat{\sigma}_\infty)$ for $\alpha=\infty$. In the plot to the right, the two curves intersect only at $(0,\widehat{\sigma}_\infty)$ for $\alpha=\infty$. The two cases of intersections are differentiated by the gap between $\widehat{\sigma}_\infty$ and $\widehat{\mu}_0$. While $\widehat{\sigma}_\infty>\widehat{\mu}_0$ always, if the samples are such that $\widehat{\sigma}_\infty$ is significantly above $\widehat{\mu}_0$, the two curves appear to be intersecting only at $\alpha=\infty$.

\begin{lemma}
    The implicit equations for joint ML estimation in Lemma \ref{lem:gmmjointml} reduce to the following equation:
    \begin{equation}
        \widehat{\mu}_\alpha = g(\alpha)\triangleq\sqrt{\widehat{\sigma}^2_\infty + \alpha^2/4} - \alpha/2.
    \label{eq:jointmlsingle}
    \end{equation}
    If there is no finite solution, the joint ML estimate is $\widehat{\mu}_{ML}=0$,  $\widehat{\sigma}^2_{ML}=\widehat{\sigma}^2_\infty$ ($\alpha\to\infty$). If there is a finite solution $\alpha^*$, the joint ML estimate is $\widehat{\mu}_{ML}=\widehat{\mu}_{\alpha^*}$, $\widehat{\sigma}^2_{ML}=\widehat{\sigma}^2_{\alpha^*}$. In either case, we have $\alpha^* \ge \alpha_l \triangleq \sigma^2_s/\widehat{\mu}_0$ and $\sigma^2_s\le\widehat{\sigma}^2_{ML}\le\widehat{\sigma}^2_\infty$.
    \label{lem:jointmlsolution}
\end{lemma}
\begin{proof}
Using $\widehat{\sigma}^2_\alpha=\alpha\widehat{\mu}_\alpha$ in $\widehat{\mu}_\alpha^2+\widehat{\sigma}_\alpha^2=\widehat{\sigma}_\infty^2$, we get $\widehat{\mu}_\alpha^2 + \alpha\widehat{\mu}_\alpha = \widehat{\sigma}_\infty^2$. Completing the square and simplifying, we obtain \eqref{eq:jointmlsingle}.

Both $\widehat{\mu}_\alpha$ and $g(\alpha)$ are decreasing in $\alpha$ with $\widehat{\mu}_0\le g(0)=\widehat{\sigma}_\infty$. Also, as $\alpha\to\infty$, we have $\widehat{\mu}_\alpha\to0$ and $g(\alpha)\to0$, which means that $\widehat{\mu}_\alpha$ and $g(\alpha)$ meet as $\alpha\to\infty$ at $(\widehat{\mu}_\alpha \to 0, \widehat{\sigma}_\alpha \to \widehat{\sigma}_\infty)$. 

For $\alpha_l=\sigma^2_s/\widehat{\mu}_0=(\widehat{\sigma}^2_\infty-\widehat{\mu}_0^2)/\widehat{\mu}_0$, we have $\widehat{\sigma}^2_\infty=\alpha_l\widehat{\mu}_0+\widehat{\mu}_0^2$ and $g(\alpha_l)=\widehat{\mu}_0$. Since we have $\mu_{\alpha}\le \widehat{\mu}_0=g(\alpha_l)$, we see that intersection of $\widehat{\mu}_\alpha$ and $g(\alpha)$ can happen only for $\alpha^*\ge\alpha_l$.

Since $\widehat{\mu}_\alpha^2 + \widehat{\sigma}_\alpha^2 = \widehat{\sigma}_\infty^2$, clearly we have $\widehat{\sigma}^2_{ML}\le\widehat{\sigma}^2_\infty$. Since $\sigma_s^2+\widehat{\mu}_0^2=\widehat{\sigma}^2_\infty$ and $\widehat{\mu}_\alpha\le\widehat{\mu}_0$, we have that $\sigma_s^2+\widehat{\mu}_\alpha^2\le\widehat{\sigma}^2_\infty$ for all $\alpha$. So, \eqref{eq:gmmcircle} can only be satisfied if $\widehat{\sigma}^2_{ML}\ge\sigma_s^2$.
\end{proof}
The upper bound $\widehat{\sigma}^2_{ML}<\widehat{\sigma}^2_\infty$ is loose when the two curves intersect at a finite $\alpha$. Tightening this bound for the joint ML estimator is a subject of current work. However, by introducing an approximation to the function $\widehat{\mu}_\alpha$, we are able to bound the variance estimate and proceed with the analysis of recursive training.

\subsection{Approximate joint ML estimation in a GMM}
\label{sec:approxjointml}
\begin{figure}[htb]
    \centering
   \conf{ \resizebox{\linewidth}{!}}   \arxiv{ \resizebox{0.7\linewidth}{!}}{\input{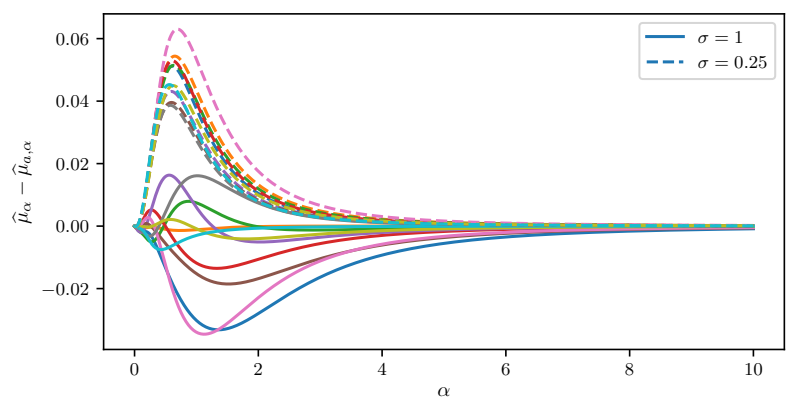}}
    \caption{Error in approximation of $\widehat{\mu}_\alpha$, $\mu=1$, $n=16$, $a=50$, 10 trials.}
    \label{fig:approx}
\end{figure}

The following upper bound for $\tanh$ with $a=2$ is from \cite{Bhayo2015}.
\begin{equation*}
    \tanh x \le u(x)=\frac{ax}{\sqrt{a^2x^2+(a+1)^2}-1}.
\end{equation*}
The upper bound $u(x)$ is fairly close to $\tanh x$ over the entire range $x>0$, and $u(0)=0$, $u(\infty)=1$ and $\lim_{x\to0} u(x)/x = 1$, which mirror the properties of $\tanh x$. In fact, it is possible to show a lower bound of a very similar form. We have that
\begin{equation*}
l(x)=\frac{ax}{\sqrt{b^2x^2+(a+1)^2}-1}\le \tanh x \le u(x)=\frac{ax}{\sqrt{a^2x^2+(a+1)^2}-1}
\end{equation*}
for a suitable value of $b>a=2$. Using a similar form, we propose the following approximation for $\widehat{\mu}_\alpha$:
\begin{equation*}
    \widehat{\mu}_\alpha=\frac{1}{n}\sum_{i=1}^n x_i\tanh \frac{x_i}{\alpha} \approx \widehat{\mu}_{a,\alpha} \triangleq \frac{a\widehat{\mu}_0^2}{\sqrt{a^2\widehat{\mu}_0^2 + \left(\dfrac{a\widehat{\mu}_0^2}{\widehat{\sigma}^2_\infty} + 1\right)^2\alpha^2} - \alpha},
\end{equation*}
where $a\ge2$ is a parameter. The approximation is carefully structured to preserve $\widehat{\mu}_{a,0} = \widehat{\mu}_0$, $\widehat{\mu}_{a,\infty} = \widehat{\mu}_\infty$ and $\lim_{\alpha\to\infty}\alpha\widehat{\mu}_{a,\alpha} = \widehat{\sigma}^2_\infty$. Fig. \ref{fig:approx} shows the error in the approximation for $n=16$ samples generated $10$ times with $\mu=1$, $\sigma=1$ and $\mu=1$, $\sigma=0.25$. 
The parameter $a$ was chosen to be $50$. We see that the maximum error is bounded within about $0.06$ over the 10 trials.

In Lemma \ref{lem:gmmjointml} and \eqref{eq:jointmlsingle}, $\widehat{\mu}_\alpha$ is replaced with its approximate version $\widehat{\mu}_{a,\alpha}$ to obtain an approximate joint ML estimator, which is the intersection of the following two curves:
\begin{align}
    (\widehat{\mu}_{a,\alpha},\widehat{\sigma}_{a,\alpha}):&\ \widehat{\mu}_{a,\alpha} = \frac{a\widehat{\mu}_0^2}{\sqrt{a^2\widehat{\mu}_0^2 + \left(\dfrac{a\widehat{\mu}_0^2}{\widehat{\sigma}^2_\infty} + 1\right)^2\alpha^2} - \alpha},\quad \widehat{\sigma}^2_{a,\alpha} = \alpha\widehat{\mu}_{a,\alpha}\text{  for  }\alpha\in(0,\infty),\label{eq:curveapprox1}\\
    \text{circle}:&\ \widehat{\mu}^2+\widehat{\sigma}^2=\widehat{\sigma}_\infty^2. \label{eq:circleapprox2}
\end{align}
Unlike the joint ML estimator, the intersection of the two curves in the above approximate version can be characterized explicitly. Let $c>0$ and $\alpha_c$ be such that the point $(\widehat{\mu}_{a,\alpha_c},\sqrt{c}\sigma_s)$ is on the curve given by \eqref{eq:curveapprox1} and, therefore, satisfying $\alpha_c\widehat{\mu}_{a,\alpha_c}=c\sigma^2_s$. Using the expression for $\widehat{\mu}_{a,\alpha_c}$ and simplifying algebraically, we obtain
\begin{equation}
\alpha^2_c(\widehat{\sigma}^2_\infty-c\sigma^2_s)(a\widehat{\mu}^2_0(\widehat{\sigma}^2_\infty+c\sigma^2_s)+2c\sigma^2_s\widehat{\sigma}^2_\infty)=ac^2\sigma^4_s\widehat{\sigma}^4_\infty.    
\label{eq:alphac}
\end{equation}
We require $c\sigma^2_s\le\widehat{\sigma}^2_\infty$ to obtain a possible intersection point. For the point $(\widehat{\mu}_{a,\alpha_c},\sqrt{c}\sigma_s)$, which can be written as $(c\sigma^2_s/\alpha_c,\sqrt{c}\sigma_s)$ to also lie on the circle $\widehat{\mu}^2+\widehat{\sigma}^2=\widehat{\sigma}^2_\infty$, we require that
$$\frac{c^2\sigma^4_s}{\alpha_c^2}+c\sigma^2_s=\widehat{\sigma}^2_\infty\text{ or }\alpha^2_c(\widehat{\sigma}^2_\infty-c\sigma^2_s)=c^2\sigma^4_s.$$
Using the above in \eqref{eq:alphac} and simplifying using $\widehat{\mu}^2_0+\sigma^2_s=\widehat{\sigma}^2_\infty$, we get
\begin{align*}
    c &= \frac{a\widehat{\sigma}^2_\infty}{a\widehat{\mu}^2_0+2\widehat{\sigma}^2_\infty},
\end{align*}
which is valid if it satisfies $c\sigma^2_s\le \widehat{\sigma}^2_\infty$. Substituting the expression for $c$ and simplifying, we get the condition
\begin{align*}
\widehat{\sigma}_\infty \le \sqrt{\frac{2a}{a-2}}\,\widehat{\mu}_0.
\end{align*}
We summarise the approximate joint ML estimator in the following lemma.
\begin{lemma}    \label{lem:jointmlapprox}
    For $a>2$, the approximate joint ML estimates $\widehat{\mu}_{a,ML}$, $\widehat{\sigma}_{a,ML}$ obtained as the intersection of the two curves in \eqref{eq:curveapprox1} and \eqref{eq:circleapprox2} are given by the following:
    \begin{equation*}
        \widehat{\sigma}^2_{a,ML}=\begin{cases}
            \dfrac{a\widehat{\sigma}^2_\infty\sigma^2_s}{a\widehat{\mu}^2_0+2\widehat{\sigma}^2_\infty}&\widehat{\sigma}_{\infty}\le\sqrt{\frac{2a}{a-2}}\,\widehat{\mu}_0,\\[15pt]
            \widehat{\sigma}^2_\infty&\text{otherwise,}
        \end{cases}\qquad \text{ and }\widehat{\mu}_{a,ML}=\sqrt{\widehat{\sigma}^2_\infty-\widehat{\sigma}^2_{a,ML}}.
    \end{equation*}
    For $0\le\kappa\le1$,  
    $$\sigma_s^2\le \widehat{\sigma}^2_{a,ML} \le (1+\kappa)\sigma_s^2\quad\text{for}\quad \frac{2\widehat{\sigma}^2_\infty}{\sigma_s^2}\le a\le A,$$
    where 
    $$A=\begin{cases}
        \infty&\widehat{\sigma}^2_\infty\le(1+\kappa)\widehat{\mu}^2_0,\\
        \dfrac{2(1+\kappa)\widehat{\sigma}^2_\infty}{\widehat{\sigma}^2_\infty-(1+\kappa)\widehat{\mu}^2_0}&\text{otherwise}.
    \end{cases}$$
\end{lemma}
\begin{proof}
The expression for $\widehat{\sigma}^2_{a,ML}$ was derived just ahead of the statement of the lemma. 

The conditions on $a$ are obtained by simplifying the lower and upper bound conditions on $\widehat{\sigma}^2_{a,ML}$.
\end{proof}

\subsection{Comparison of ML and approximate ML estimators}
\label{sec:gmmcomparison}

\begin{figure}[htp]

\begin{subfigure}{\textwidth}
\centering
\resizebox{0.7\linewidth}{!}{\input{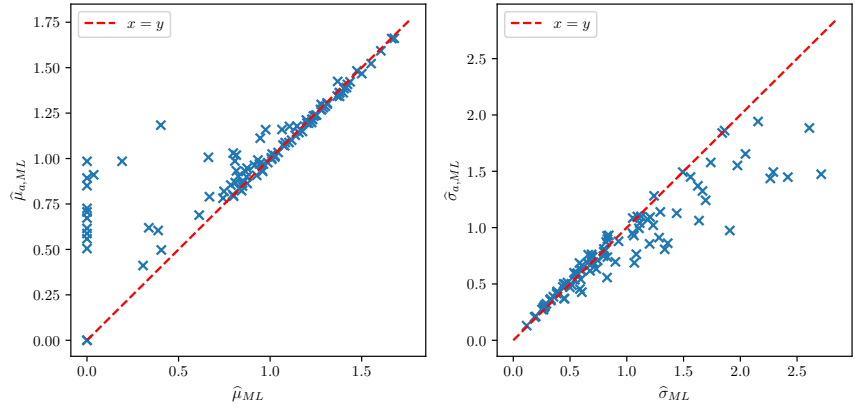}}
\caption{$\mu=1,\sigma=1$, $a=50$, $n=10$}
\end{subfigure}

\bigskip

\begin{subfigure}{\textwidth}
\centering
\resizebox{0.8\linewidth}{!}{\input{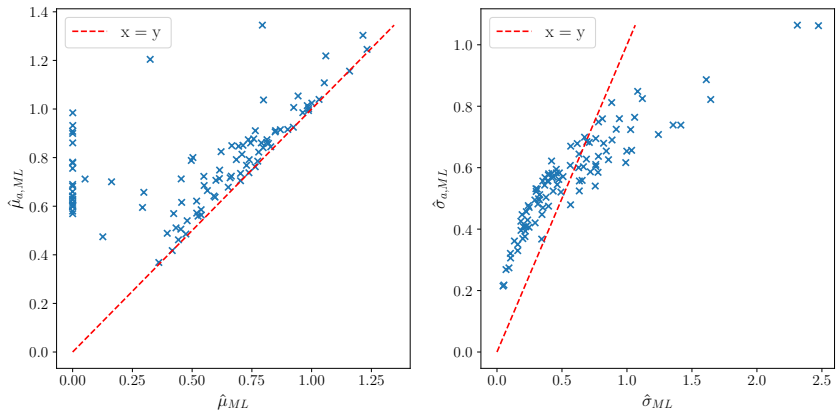}}
\caption{$\mu=0.1,\sigma=1$, $a=50$, $n=10$}
\end{subfigure}

\bigskip

\begin{subfigure}{\textwidth}
\centering
\resizebox{0.8\linewidth}{!}{\input{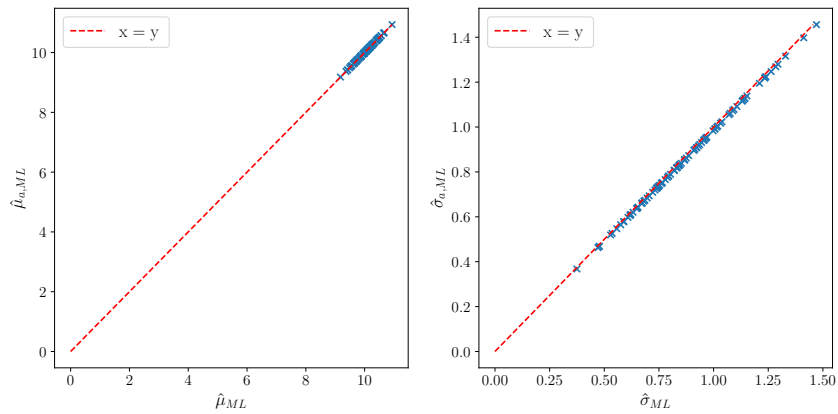}}
\caption{$\mu=10,\sigma=1$, $a=50$, $n=10$}
\end{subfigure}

\caption{Comparison of $100$ runs of joint ML and approximate joint ML estimation for GMM with various $\mu$.}
\label{fig:gmmscatter}
\end{figure}

Figure \ref{fig:gmmscatter} shows a scatter plot of estimated means and variances under the joint ML estimator and the approximate one for various values of $\mu_0$. We observe that the scatter plot is close to the $x=y$ line showing that the approximation is quite accurate in most of the cases. 

\subsection{Model collapse: Estimator and Proof of Theorem \ref{thm:gmm}}
\label{sec:gmmthmproof}
Algorithm \ref{alg:approx} provides the pseudocode for the approximate joint ML estimator that is used in studying model collapse for the GMM recursive training process in each round of training.

\begin{algorithm}[hbt]
    \caption{Approximate joint ML estimator}\label{alg:approx}
    \begin{algorithmic}[1]
        \Require Integer $n>3$, Samples $x_1,\ldots,x_n$, $\kappa:\log(1+\kappa)=\frac{1}{2n(4n-1)}$
        \State $\widehat{\mu}_0\gets\frac{1}{n}\sum_{i=1}^n|x_i|$, $\widehat{\sigma}^2_\infty\gets\frac{1}{n}\sum_{i=1}^nx^2_i$, $\sigma^2_s\gets\widehat{\sigma}^2_\infty-\widehat{\mu}^2_0$
        \If{$\widehat{\sigma}^2_\infty\le(1+\kappa)\widehat{\mu}^2_0$}
            \State $a\gets50$ \Comment{can be any other high value}
        \Else
            \State $a\gets\dfrac{2(1+\kappa)\widehat{\sigma}^2_\infty}{\widehat{\sigma}^2_\infty-(1+\kappa)\widehat{\mu}^2_0}$
        \EndIf
        \If{$\widehat{\sigma}^2_\infty\le\frac{2a}{a-2}\widehat{\mu}^2_0$}
            \State $\widehat{\sigma}^2_{a,ML} \gets \dfrac{a\widehat{\sigma}^2_\infty\sigma^2_s}{a\widehat{\mu}^2_0+2\widehat{\sigma}^2_\infty}$
        \Else
            \State $\widehat{\sigma}^2_{a,ML} \gets \widehat{\sigma}^2_\infty$
        \EndIf
        \State $\widehat{\mu}_{a,ML}=\sqrt{\widehat{\sigma}^2_\infty-\widehat{\sigma}^2_{a,ML}}$
    \end{algorithmic}
\end{algorithm}
The proof for the collapse of the variance proceeds as follows. Using $\Sigma^2_k\le(1+\kappa)\Sigma^2_{k-1}\frac{\chi^2_{n-1}}{n-1}$ and borrowing from the analysis for the Gaussian case, we have that (recall that $m=n-1$)
\begin{align}
E[\Sigma_k]&\le (1+\kappa)^{1/2}E[\Sigma_{k-1}]\exp\left\{-\frac{1}{4n-1}\right\}\nonumber\\
&=E[\Sigma_{k-1}]\exp\left\{-\frac{1}{4n-1}+\frac{1}{2}\log(1+\kappa)\right\}\nonumber\\
&=E[\Sigma_{k-1}]\exp\left\{-\frac{1}{4n}\right\},\nonumber
\end{align}
where we choose $\kappa$ such that $\frac{1}{2}\log(1+\kappa)=\frac{1}{4n(4n-1)}$.
Using the above bound repeatedly,
$$E[\Sigma_k]\le \sigma_0\exp\left\{-\frac{k}{4n}\right\}.$$
Use of Markov's inequality as in the Gaussian case completes the proof.

\section{Additional experiments} \label{sec:additional_experiments}

\begin{figure*}[htb]
    \centering
    \begin{subfigure}{0.5\textwidth}
        \centering
         \resizebox{0.9\linewidth}{!}{\input{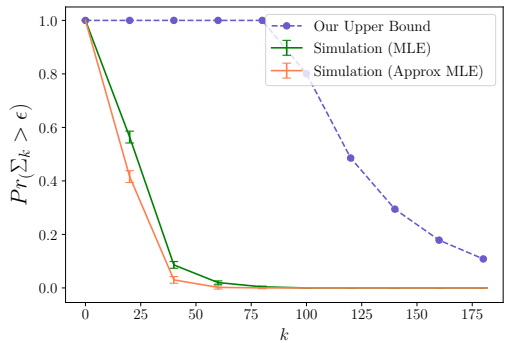}}
        \caption{$\mu_0 = 10$, $\sigma_0 = 1$}
    \end{subfigure}%
    \hfill
    \begin{subfigure}{0.5\textwidth} 
        \centering
        \resizebox{0.9\linewidth}{!}{\input{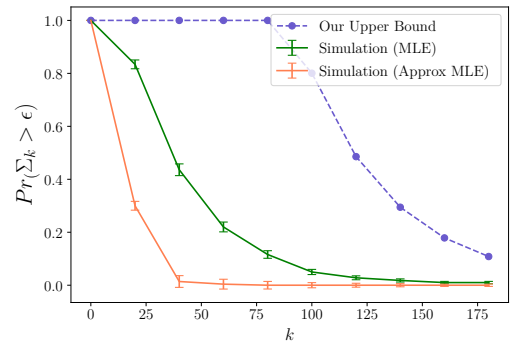}}
        \caption{$\mu_0 = 0.1$, $\sigma_0 = 1$}
    \end{subfigure}
    \caption{Comparison of theoretical upper bound with empirical observations of  $\text{Pr}(\Sigma_k>\epsilon)$ in GMM recursive training process $(\epsilon = 10^{-1}, n = 15)$ with Joint ML and Approximate Joint ML estimators.}
    \label{fig:gmm_collapse_prob_varied_mu}
 \end{figure*}

In Figure~\ref{fig:gmm_collapse_prob_varied_mu}, we show $\text{Pr}(\Sigma_k>\epsilon)$ versus $k$ in GMM for the joint ML estimator and the approximate ML estimator and the upper bound from Theorem \ref{thm:gmm}.

\label{app:experiments}

\end{document}